\documentclass{ieeeaccess}

\usepackage{cite}
\usepackage{amsmath,amssymb,amsfonts}
\usepackage{algorithmic}
\usepackage{graphicx}
\usepackage{textcomp}
\usepackage{fancybox}
\usepackage{algorithm}
\usepackage{array}
\usepackage[caption=false,font=scriptsize,labelfont=sf,textfont=sf]{subfig}
\usepackage{stfloats}
\usepackage{url}
\usepackage{verbatim}
\usepackage{pifont}
\usepackage{wasysym}
\usepackage{hyperref}
\hypersetup{colorlinks=true, linkcolor=blue, anchorcolor=red, citecolor=blue, filecolor=red, urlcolor=blue, pdfauthor=author}

\usepackage{graphicx}
\usepackage{subfig}

\usepackage{booktabs}
\usepackage{multirow}
\usepackage{makecell}
\usepackage{textcomp}

\usepackage{enumitem} 
\usepackage{placeins}
\usepackage{afterpage}

\usepackage{setspace}

\def\BibTeX{{\rm B\kern-.05em{\sc i\kern-.025em b}\kern-.08em
    T\kern-.1667em\lower.7ex\hbox{E}\kern-.125emX}}
\begin{document}
\history{Date of publication xxxx 00, 0000, date of current version xxxx 00, 0000.}
\doi{10.1109/ACCESS.2017.DOI}

\title{
From Pixels to Images: A Structural Survey of Deep Learning Paradigms in Remote Sensing Image Semantic Segmentation
}

\author{
\uppercase{Quanwei Liu}\authorrefmark{1,2}, \IEEEmembership{Student Member, IEEE},
\uppercase{Tao Huang} \authorrefmark{1,2}, \IEEEmembership{Senior Member, IEEE},
\uppercase{Jiaqi Yang} \authorrefmark{3} \IEEEmembership{Member, IEEE},
\uppercase{Wei Xiang} \authorrefmark{4}, \IEEEmembership{Senior Member, IEEE}}

\address[1]{College of Science and Engineering, James Cook University, Cairns QLD 4878, Australia (e-mail: quanwei.liu@my.jcu.edu.au, tao.huang1@jcu.edu.au)}
\address[2]{Center for AI and Data Science Innovation, James Cook University, Cairns QLD 4878, Australia}
\address[3]{Department of Forest and Wildlife Ecology, University of Wisconsin-Madison, Madison, WI 53706 USA (e-mail: jiaqi.yang@wisc.edu)}
\address[4]{School of Computing, Engineering and Mathematical Sciences, La Trobe University, Melbourne, VIC 3086, Australia (e-mail: W.Xiang@latrobe.edu.au)}
 
\tfootnote{This work was supported in part by the Australian Government through the Australian Research Council's Discovery Projects Funding Scheme under Project DP220101634.}

\markboth
{Author \headeretal: Preparation of Papers for IEEE TRANSACTIONS and JOURNALS}
{Author \headeretal: Preparation of Papers for IEEE TRANSACTIONS and JOURNALS}

\corresp{Corresponding author: Tao Huang (e-mail: tao.huang1@jcu.edu.au).}

\begin{abstract}

Remote sensing images (RSIs) capture both natural and human-induced changes on the Earth's surface, serving as essential data for environmental monitoring, urban planning, and resource management.
Semantic segmentation (SS) of RSIs enables the fine-grained interpretation of surface features, making it a critical task in RS analysis.
With the increasing diversity and volume of RSIs collected by sensors on various platforms, traditional processing methods struggle to maintain efficiency and accuracy.
In response, deep learning (DL) has emerged as a transformative approach, enabling substantial advances in remote sensing image semantic segmentation (RSISS) by automating hierarchical feature extraction and high segmentation performance.
As researchers continue to explore end-to-end SS, DL-based RSISS has undergone a structural evolution from pixel-level and patch-based classification to tile-level and image-level segmentation.
However, existing reviews often focus on individual components, such as supervision strategies or fusion stages, and lack a unified operational perspective aligned with segmentation granularity and the training/inference pipeline.
This paper provides a comprehensive review by organizing DL-based RSISS into a pixel–patch–tile–image hierarchy, covering early pixel-based methods, prevailing patch-based and tile-based techniques, and emerging image-based approaches.
Specifically, the survey analyzes four supervision strategies, eleven feature extraction strategies, and six information fusion strategies, revealing the field’s progression from local to global feature extraction, from traditional DL architectures to foundation models, and from unimodal to multimodal segmentation.
This review offers a holistic and structured understanding of DL-based RSISS, highlighting representative datasets, comparative insights, and open challenges related to data scale, model efficiency, domain robustness, and multimodal integration.
Furthermore, to facilitate reproducible research, curated code collections are provided at: \href{https://github.com/quanweiliu/RSISS}{RSISS}.

%
\end{abstract}


\titlepgskip=-15pt

\maketitle

\section{Introduction}
\label{Introduction}

\Figure[t](topskip=0pt, botskip=0pt, midskip=0pt)
[width=\linewidth]{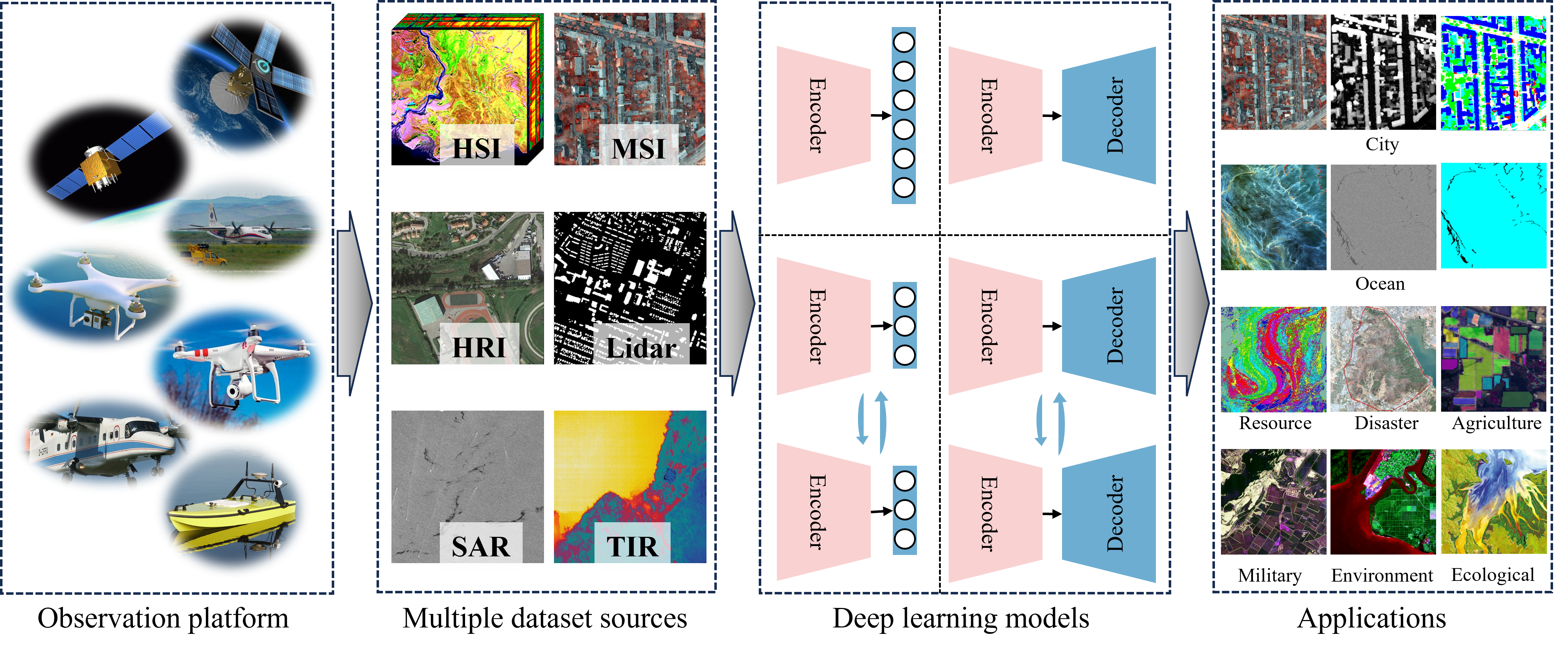}
{\textbf{Processing flow for remote sensing image semantic segmentation.} $\downarrow\uparrow$ \textbf{denotes the feature interaction.}\label{semanticSegmentationFramework}}

Diverse perception modalities have emerged alongside advances in electromagnetic spectrum research, including remote sensing (RS) via satellites, aircraft, and unmanned aerial vehicles (UAVs) equipped with various sensors.
These modalities exhibit distinct spatial, spectral, temporal, and radiometric resolution characteristics. For instance, hyperspectral images (HSIs) offer both spatial context and detailed spectral information across numerous bands, while high-resolution imagery (HRI) provides finer textural detail and enhances the discrimination of small targets.
Remote sensing image semantic segmentation (RSISS) seeks to classify RS imagery into distinct categories on a pixel-wise basis, enabling fine-grained interpretation of the Earth’s surface.
This capability plays a critical role in marine \cite{KIKAKI202439}, urban \cite{ feng2019multisource}, forest \cite{chen2023semi}, arable \cite{qu2023vegetation}, and disaster-related applications \cite{casagli2023landslide}. The semantic segmentation (SS) process is illustrated in Figure~\ref{semanticSegmentationFramework}.

Early RSISS methods were primarily based on machine learning (ML) techniques relying on handcrafted feature extraction, including texture, structural, spectral, and scattering descriptors, combined with classifiers such as support vector machines (SVMs) and random forests.
Although these approaches achieved notable progress, performance improvements often required increasingly complex feature engineering, limiting scalability and generalization \cite{yu2023deep}.
The emergence of deep learning (DL) fundamentally transformed this paradigm by enabling automatic hierarchical feature extraction through data-driven architectures.

As researchers continue to explore end-to-end SS, DL-based RSISS has undergone a structural evolution in how segmentation is performed.
Early DL models operated primarily at the pixel or patch level, where each pixel or local region was processed independently.
In such settings, images were divided into numerous pixels or patches, and predictions were aggregated to form segmentation maps. Such methods are commonly referred to as classification models, specifically the pixel- or patch-level classification model. 
While pixel- and patch-based approaches require relatively small training data and model parameters, they suffer from repeated computations and limited global contextual awareness, which constrain generalization capability.
As larger public datasets with pixel-level annotations became available, fully convolutional networks (FCNs) \cite{long2015fully}, U-Net \cite{ronneberger2015u}, vision foundation models (VFMs) \cite{wang2023samrs} and computer cluster \cite{chen2025rest} enabled tile- and image-level semantic segmentation, significantly improving computational efficiency and contextual modeling.

\begin{figure}[!t]
    \centering
        \includegraphics[scale=0.45]{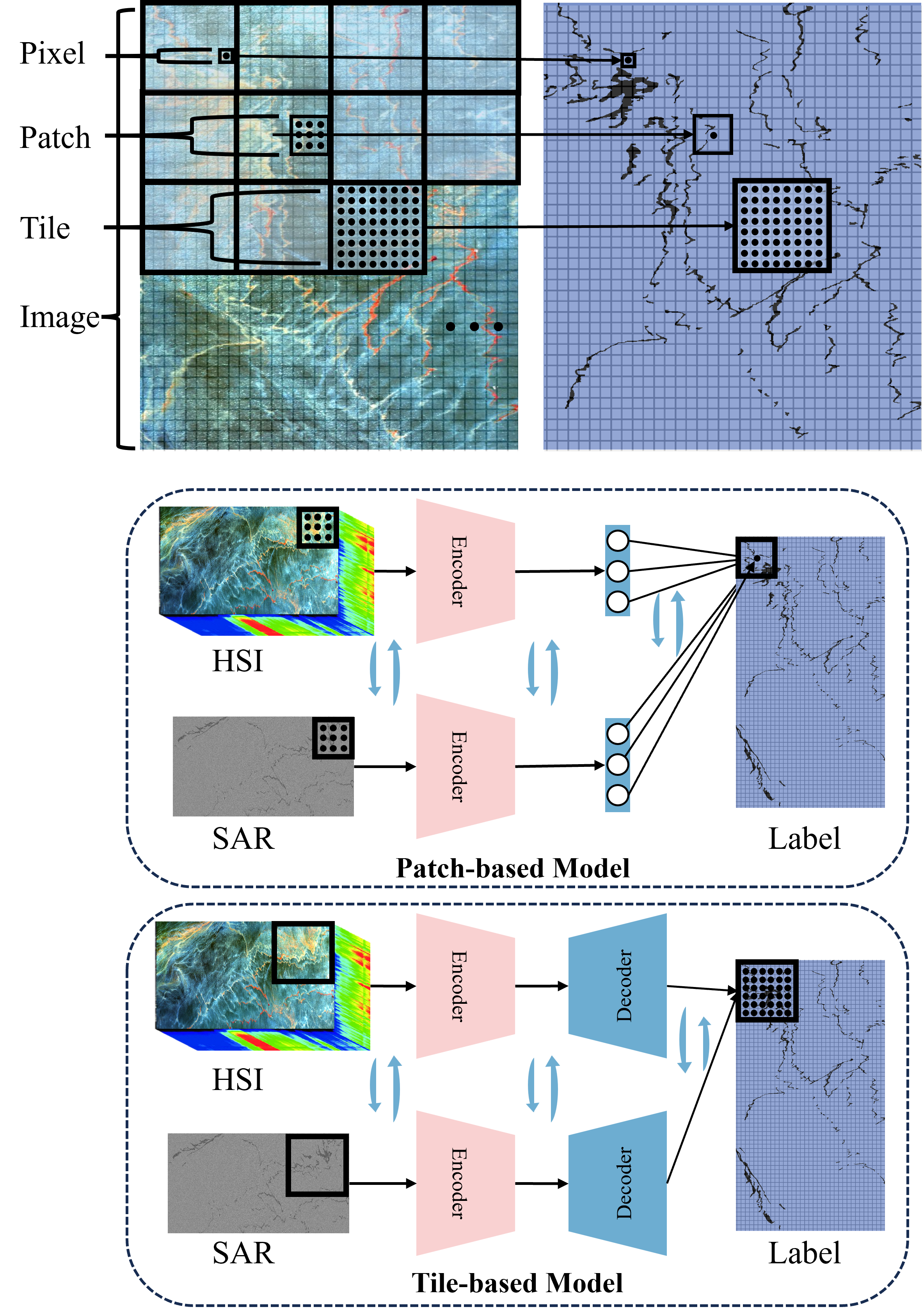}
        \caption{\textbf{Illustration of the proposed pixel–patch–tile–image taxonomy for RIS classification and semantic segmentation.
        Pixel-based methods perform per-pixel classification, predicting a class label for each pixel independently. 
        Patch-based classification assigns one label to each patch, where a local neighbourhood centred at a target pixel is used as input to predict a single class, and the prediction is then mapped back to the patch centre. 
        Tile-based semantic segmentation processes a larger image tile in one forward pass and outputs a dense segmentation map for all pixels within the tile simultaneously. 
        Image-based segmentation generalizes tile-based inference to arbitrary image sizes and resolutions, producing consistent dense predictions for large-area RSIs.
        }}
    \label{ImageTilePatch}
    \vspace{-1.0em}
\end{figure}

This evolution from pixel-based classification to image-level segmentation reflects not only architectural innovation but also the co-development of data scale and feature extraction mechanisms.
Across these regimes, advances in feature extraction, such as multi-scale modeling, global context learning, representation transfer, and domain-robust training, have played a central role in expanding receptive fields, improving robustness, and enabling cross-domain generalization.
Accordingly, this paper organizes DL-based RSISS into a unified pixel–patch–tile–image hierarchy, explicitly linking segmentation granularity to training/inference pipelines and feature extraction strategies, as illustrated in Figure~\ref{ImageTilePatch}.

Beyond unimodal feature extraction, multimodal data fusion further enriches semantic representation.
Different sensing modalities provide complementary information: for example, concrete pavements and roofs may share similar spectral signatures in HSI but differ in elevation captured by LiDAR \cite{xue2022deep}, while certain vegetation types may be indistinguishable in LiDAR yet separable in HSI.
Fusion of modalities such as HSI, LiDAR, multispectral imagery, and SAR enables more comprehensive and robust scene understanding.
In this survey, multimodal fusion is examined as a complementary dimension that interacts with the proposed pixel–patch–tile–image framework.


Table~\ref{survey} explicitly compares our proposed framework with existing surveys. Although numerous surveys have reviewed remote sensing image semantic segmentation from perspectives such as spectral and spatial feature extraction, supervision strategies, and fusion stages, most focus on individual components in isolation. These conventional taxonomies often detach algorithmic design from the realities of practical deployment.
To address this gap, this paper introduces a unified hierarchy of pixels, patches, tiles, and images. Compared to existing survey frameworks, our proposed taxonomy offers three concrete advantages:

\begin{itemize}[topsep=1pt, itemsep=2pt, parsep=0pt, partopsep=0pt]
\item It ensures direct mapping to the data processing pipeline and hardware memory constraints, providing a highly practical engineering guide for model deployment across varying dataset scales.
\item It establishes a clear historical and structural continuum, seamlessly bridging the transition from early patch classification to modern tile segmentation and emerging holistic image segmentation.
\item It enables a holistic evaluation of feature extraction and multimodal fusion strategies across different spatial granularities.
\end{itemize}

By explicitly categorizing existing literature along this structural continuum, this survey reveals underexplored regimes and offers a unified roadmap that remains insufficiently addressed in current literature.
The key contributions are as follows:
\begin{itemize}[topsep=1pt, itemsep=2pt, parsep=0pt, partopsep=0pt]
\item Proposing a unified pixel–patch–tile–image taxonomy that directly corresponds to segmentation granularity and the training/inference pipeline, clarifying how feature extraction evolves with increasing data scale and model capacity.
\item Bridging patchwise classification and tilewise segmentation through an integrated analysis of feature extraction mechanisms, supervision paradigms, and multimodal fusion strategies, revealing design trade-offs and underexplored regimes.
\item Providing practical resources for reproducible research by summarizing representative RSISS datasets and curating code collections for both patchwise and tilewise pipelines.
\end{itemize}

The remainder of this paper is organized as follows.
Section~\ref{RSISS} introduces the pixel–patch–tile–image taxonomy and categorizes existing RSISS methods according to segmentation granularity.
Section~\ref{FeatureUniExtraction} analyzes feature extraction strategies across these regimes.
Section~\ref{ImageFusion} examines multimodal fusion mechanisms within this unified framework.
Section~\ref{Datasets} lists an array of the most representative datasets for RSISS.
Section~\ref{OIFD} provides a critical reflection on current methodologies and highlights future trajectories in the field of RSISS.
Finally, Section~\ref{Conclusion} concludes the paper.

\begin{table*}[!t]
\caption{\textbf{A summary of the recently published surveys in remote sensing image semantic segmentation. The subscript in the first column indicates the year of publication. For example, Imani et al.}$_{20}$ \textbf{\cite{imani2020overview} }\textbf{indicate that this paper was published in 2020 by  Imani et al.}}
\label{survey}
\centering
\setlength\tabcolsep{3pt}
\scriptsize
\begin{tabular}{llp{3.6cm}cccccccccccccc}
\toprule
\multirow{3}{*}{Paper} &
\multirow{3}{*}{Publication} &
\multirow{3}{*}{Survey Topic} &
\multirow{3}{*}{Pix} &
\multicolumn{2}{c}{PatchU} &
\multicolumn{2}{c}{TileU} &
\multicolumn{2}{c}{PatchM} &
\multicolumn{5}{c}{TileM} &
\multirow{3}{*}{Data} \\
\cmidrule(lr){5-6}
\cmidrule(lr){7-8}
\cmidrule(lr){11-15}
\cmidrule(lr){9-10}

& & & &
\multirow{2}{*}{TDL} & \multirow{2}{*}{FM} &
\multirow{2}{*}{TDL} & \multirow{2}{*}{FM} &
\multirow{2}{*}{\makecell{Linear \\ fusion}} & \multirow{2}{*}{\makecell{Nonlinear \\ fusion}} &
\multirow{2}{*}{\makecell{Linear \\ fusion}} & \multicolumn{4}{c}{\makecell{Nonlinear fusion}} & \\
\cmidrule(lr){12-15}

& & & 
& & 
& & 
& & 
& & Att & Gat & Re & Align & 
\\
\midrule

Imani et al.$_{20}$ \cite{imani2020overview} & Inf. Fusion & Spectral and spatial fusion for hyperspectral image classification & \checked & \checked  & $\times$ & $\times$ & $\times$  & $\times$ & $\times$ & $\times$ & $\times$ & $\times$ & $\times$ & $\times$ & $\times$ \\
Rasti et al.$_{20}$ \cite{rasti2020feature} & IEEE GRSM & Feature extraction for hyperspectral imagery from Shallow to Deep& \checked & \checked &  $\times$ & $\times$ &  $\times$ & $\times$  & $\times$ & $\times$  & $\times$ & $\times$ & $\times$ &  $\times$ & $\times$ \\
Zang et al.$_{21}$ \cite{zang2021land} & IEEE JSTAR & Land-Use mapping for high-spatial resolution remote sensing image & \checked & \checked &  $\times$ & \checked & $\times$ & $\times$ & $\times$ & $\times$ &  $\times$ & $\times$ & $\times$ & $\times$ & \checked \\
Yuan et al.$_{21}$ \cite{yuan2021review} & Expert Syst. Appl. & Deep learning methods for semantic segmentation of remote sensing imagery & \checked & \checked  & $\times$ & \checked & $\times$  & $\times$ & $\times$ & \checked & $\times$ & \checked & $\times$ & $\times$ & \checked \\
Ahmad et al.$_{22}$ \cite{ahmad2021hyperspectral} & IEEE JSTAR & Hyperspectral image classification—traditional to deep models & \checked & \checked  & $\times$ & $\times$ & $\times$  & $\times$ & $\times$ & $\times$ & $\times$ & $\times$ & $\times$ & $\times$ & $\times$ \\
Yu et al.$_{23}$ \cite{yu2023deep} & Remote Sens. & Deep learning methods for semantic segmentation in remote sensing with small data & $\times$  &  $\times$  & $\times$ & \checked & $\times$  & \checked & $\times$ & \checked & $\times$ & \checked & $\times$ & $\times$ & $\times$ \\
Huang et al.$_{24}$ \cite{huang2023deep} & IEEE JSTAR & Deep-learning-based semantic segmentation of remote sensing images & $\times$ & $\times$  & $\times$ & \checked & \checked & $\times$ & $\times$  & \checked & \checked & \checked & \checked & $\times$ & \checked \\
Liu et al.$_{24}$ \cite{liu2024review} & IEEE JSTAR & Optical and SAR image deep feature fusion in semantic segmentation & \checked & \checked  & $\times$ & \checked & \checked  & $\times$ & $\times$ & \checked & \checked & \checked & \checked & \checked & \checked \\
Kumar et al.$_{24}$ \cite{kumar2024deep} & Comput. Sci. Rev. &  Deep learning for hyperspectral image classification & \checked & \checked  & $\times$ & $\times$ & $\times$ & \checked & $\times$ & $\times$ & $\times$ & $\times$ & $\times$ & $\times$ & \checked \\
Akewar et al.$_{25}$ \cite{akewar2024integration} & IEEE GRSM &  An integration of natural language and hyperspectral imaging & \checked & \checked & \checked & \checked & \checked & $\times$ & $\times$ & $\times$ & $\times$ & $\times$ & $\times$ & $\times$ & $\times$ \\
Rehman et al.$_{25}$ \cite{rehman2025deep} & Artif. Intell. Rev. &  Deep learning for HS-LiDAR imagery classification & \checked & \checked  & $\times$ & $\times$ & $\times$ & \checked & $\times$ & $\times$ & $\times$ & $\times$ & $\times$ & $\times$ & \checked \\
\textbf{This Survey}  & -- & Deep learning advances in remote sensing semantic segmentation & \checked & \checked & \checked & \checked & \checked &  \checked & \checked  & \checked  & \checked &  \checked & \checked & \checked  & \checked \\
\midrule
\multicolumn{17}{p{0.98\textwidth}}{{Pix: Pixel Classification, PatchU: Patchwise Unimodality Classification, TileU: Tilewise Unimodal Segmentation, PatchM: Patchwise Multi-modality Classification, TileM: Tilewise Multi-modality Segmentation, TDL: Traditional DL Model, FM: Foundation Model, Att: Attention Mechanism Fusion, Gat: Gated Mechanism Fusion, Re: Reconstruction Mechanism Fusion, Align: Alignment Mechanism Fusion.} \par
\textit{Journal abbreviations:} 
Inf. Fusion: \textit{Information Fusion}; 
IEEE GRSM: \textit{IEEE Geoscience and Remote Sensing Magazine}; 
IEEE JSTAR: \textit{IEEE Journal of Selected Topics in Applied Earth Observations and Remote Sensing}; 
Remote Sens.: \textit{Remote Sensing}; 
Expert Syst. Appl.: \textit{Expert Systems with Applications}; 
Comput. Sci. Rev.: \textit{Computer Science Review}; 
Artif. Intell. Rev.: \textit{Artificial Intelligence Review}.
} \\
\bottomrule
\end{tabular}
\end{table*}

\begin{figure}[t]
    \centering
        \includegraphics[scale=0.5]{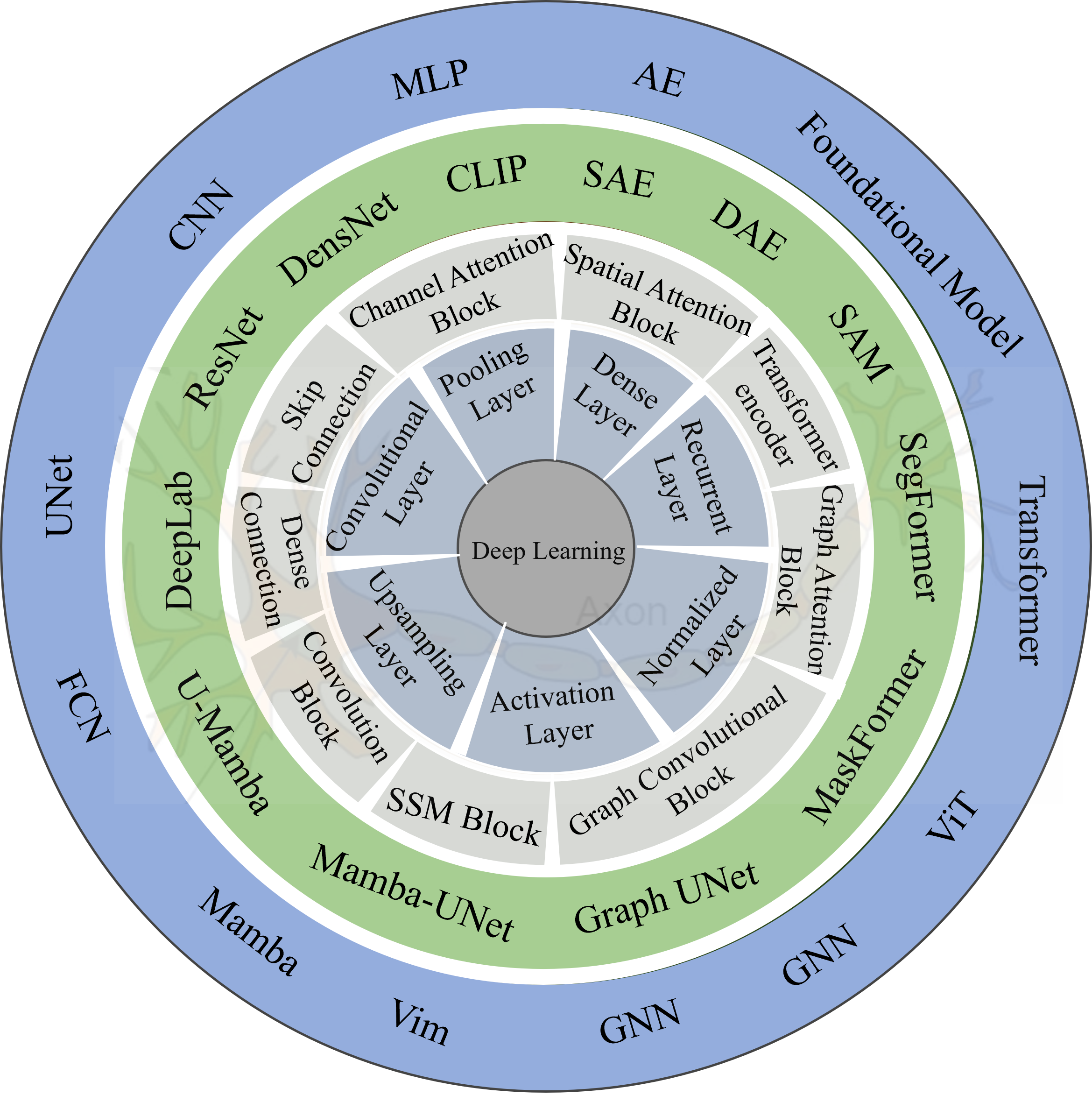}
        \caption{\textbf{Configurable architecture for DL models. CNN: convolutional neural network; MLP: multilayer perceptron; AE: autoencoder; SAE: stacked autoencoder; DAE: denoising autoencoders; SAM: Segment Anything Model; CLIP: contrastive language-image pretraining; FCN: fully convolutional network; Vim: Vision Mamba; ViT: Vision Transformer.}}   
    \label{framework_deep_network}
\end{figure}

\section{Remote sensing image semantic segmentation }
\label{RSISS}

Since the success of AlexNet at the 2012 ILSVRC \cite{krizhevsky2012imagenet}, DL has rapidly attracted attention. A series of convolutional neural network (CNN) variants have been designed to replace traditional statistical learning and ML algorithms in many domains. 
Meanwhile, a range of novel neural network architectures has been proposed to enhance the capabilities of CNN. Neural network architectures originate from permutations of neurons, as shown in Figure \ref{framework_deep_network}. 
Based on structural characteristics, DL models can be categorised into four classes following a layer–block–network–architecture framework.

Based on these DL frameworks, RSI data processing has shifted from shallow, hand-crafted feature extraction to deep feature extraction.
This paper categorizes DL-based SS approaches into pixel- and patch-based classification, and tile- and image-based segmentation methods, as illustrated in Figure \ref{ImageTilePatch}. 
Compared with conventional taxonomies based on spectral–spatial cues, supervision types, or fusion stages, this pixel–patch–tile–image hierarchy offers a more operational view that is tightly coupled to the feature extraction and training/inference pipeline. 
It accommodates both unimodal and multimodal settings and aligns naturally with the historical evolution from early MLP to current foundation models. 
This makes it easier to systematically compare architectures, data requirements, and generalization behaviours, and to reveal under-explored regimes along this progression.

\begin{table*}[ht]
\setlength{\tabcolsep}{2pt}
\caption{\textbf{Comparison of key features across different segmentation paradigms}}
\label{GranularityParadigm}
\centering
\footnotesize
\begin{tabular}{lllllll}
\toprule
\makecell[l]{Granularity \\ Paradigm} & \makecell[l]{Representative \\ Algorithms}   & \makecell[l]{Computational \\ Complexity}   & \makecell[l]{Parameter Count} & Dataset Type / Scale & Supervision Requirement & Generalization Capability \\
\hline
Pixel-level & MLP, SAE & Very Low & Minimal (\textless 1M) & Hyperspectral / Small & Dense pixel labels & Weak (Prone to local shifts) \\
Patch-level & 2D/3D-CNN & \makecell[l]{Low (Train) \\ High (Infer)}  & Light (0.1M - 10M) & HSI, LiDAR / Medium & Central-pixel label & Limited (Lacks global context) \\
Tile-level & U-Net, DeepLab & Moderate to High & Heavy (10M - 200M) & Optical, SAR / Large & Dense pixel/Weak labels & Strong (Within domain) \\
Image-level & REST & Extremely High & Massive (\textgreater 100M) & Global scale   & Dense pixel/Weak labels & Strong (Within domain)  \\
\bottomrule
\end{tabular}%
\end{table*}

\begin{figure*}[ht]
    \centering
    \includegraphics[width=\textwidth,origin=c]{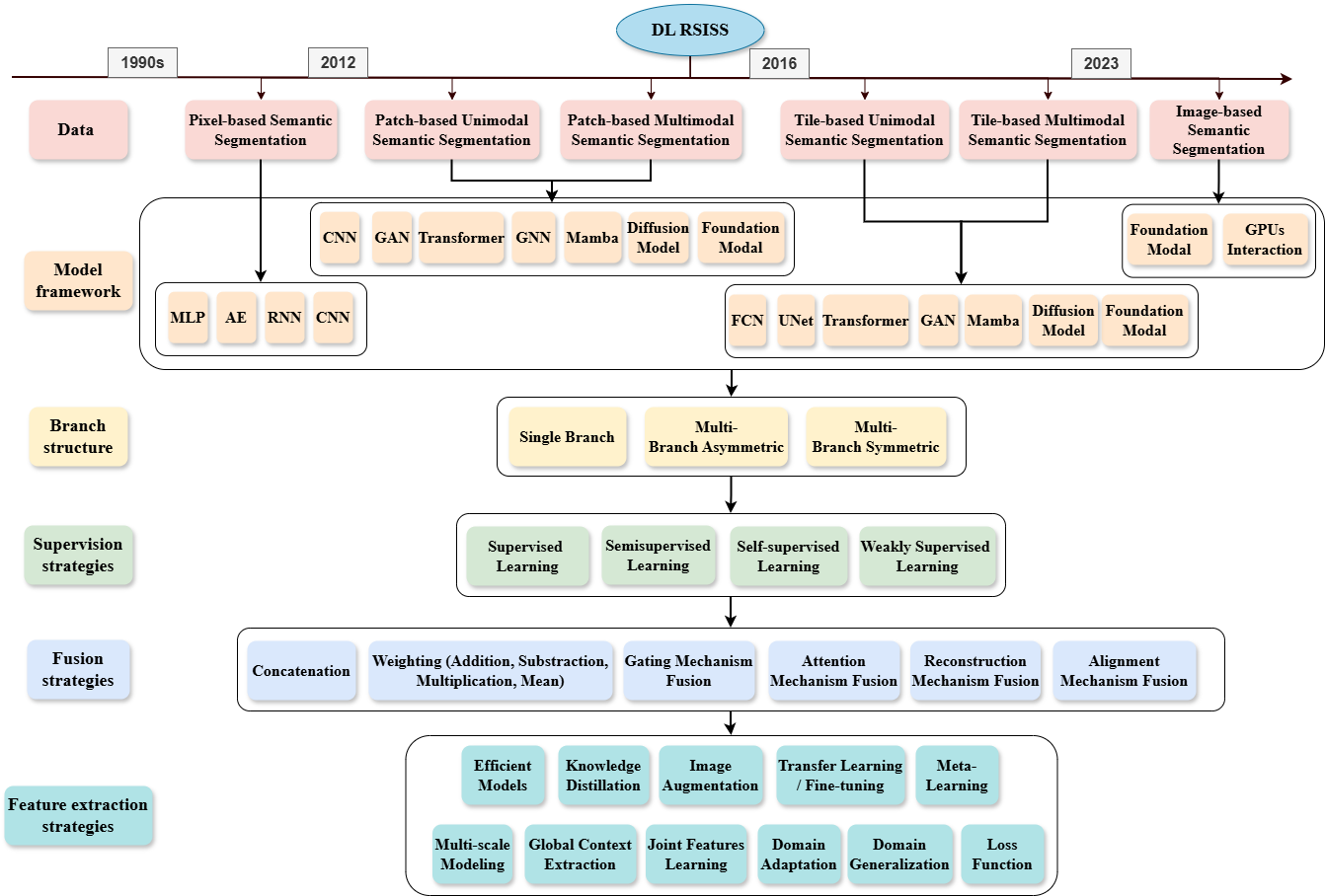}
    \caption{\textbf{Illustration of the data processing, model framework, branch structure, supervision, fusion, and feature extraction strategies used for remote sensing image semantic segmentation.}}
    \label{surveyFramework}
\end{figure*}

\subsection{Pixel-based classification}

Pixel-based classification extracts image features at the pixel level and uses them to predict target categories on a per-pixel basis.
Since the era of statistical learning, pixel-based RSISS methods have attracted increasing research attention.
In the early stage of neural network development, researchers explored the use of MLP \cite{ topouzelis2007detection}, AE \cite{deng2018active}, and RNN \cite{sun2019using} in RSISS.
For example, Topouzelis \cite{topouzelis2007detection} employed two MLPs sequentially to detect dark formations and identify oil spills or look-alike phenomena.
Chen et al. \cite{chen2014deep} introduced DL-based feature extraction for HSI classification, utilizing SAE to extract deep features in an unsupervised manner.

With the introduction of AlexNet, deep learning methods represented by convolutional neural networks (CNNs) officially entered the public consciousness \cite{krizhevsky2012imagenet}.
In \cite{hu2015deep}, Hu et al. were the first to employ CNNs with multiple layers in the spectral domain for HSI classification.
In addition, RNNs are frequently used to capture temporal dependencies in image time series, improving classification accuracy and reducing model complexity in tasks such as crop classification, where seasonal variations are significant \cite{chen2022joint}.

Pixel-based classification methods marked the introduction of DL into RSI-related computer vision tasks.
Although these methods have been surpassed by more recent models in information extraction, training efficiency, and accuracy, they have opened a new direction for future research, highlighting the significant potential of DNNs in RSI analysis.

\subsection{Patch-based classification}

Patch-based classification extracts image features from patches and uses them to predict the central pixel of each patch on a per-pixel basis. A patch refers to a small region surrounding the central pixel.
The successful application of DL models such as CNN, Transformer, and Mamba in image classification has led to the development of various patch-based variants for RSISS \cite{li2024review}.
CNNs were introduced into road extraction tasks in \cite{mnih2012learning}, where large image patches were used to provide contextual information for small-patch predictions.

To further enhance patch-based classification capability, there are two main directions. 
The first is to mine more discriminative features from existing data, and the second is to integrate richer information via fusion.
To fully exploit the information contained in the data, feature extraction strategies such as multi-scale modeling, joint feature learning, image augmentation, and domain adaptation (DA) have been developed. 
These methods explore feature representations from different perspectives and improve classification performance across diverse scenarios.
To fuse more features, multimodal fusion strategies, such as attention and gating mechanisms, are employed to regularize and align features across modalities. 
They aim to obtain more complete object representations and enhance the classification ability for specific targets.

Although these models outperformed traditional approaches, patch-based classification only predicts one label per patch. This leads to two problems: i) the inference process is inefficient. To perform global predictions, each pixel is computed multiple times; ii) the generalization capacity is weak. The limited contextual information in a small patch restricts model performance and weakens generalization capacity.
As a result, this approach is particularly suitable for scenarios involving small datasets where individual pixels carry substantial discriminative information, such as hyperspectral or multispectral tasks.

\subsection{Tile-based semantic segmentation}

Tile-based segmentation aims to extract image features by tile and utilize the extracted features to predict the segmentation map end-to-end.
Tile-based RSISS methods, primarily built upon FCN and U-Net architectures, have gained momentum due to the increasing availability of publicly labelled datasets.
These models achieve a balance between efficiency and generalization by reducing the number of trainable parameters without compromising segmentation performance \cite{kampffmeyer2016semantic, sherrah2016fully, volpi2016dense}.

Unlike patch-based methods, which infer pixel labels using patches, tile-based segmentation performs scale-invariant, pixel-level prediction.
In this approach, an input tile generates a corresponding output tile, substantially improving computational efficiency and making tile-based segmentation the foundational framework for modern SS methods.
Despite these advantages, this paradigm still faces several challenges, including increased computational burden as tile size grows and limited perceptual capability for fine-grained targets.
To overcome these challenges and improve model performance, knowledge distillation, fine-tuning, domain generalization, and multimodal data fusion methods have been advanced. 
Each of these approaches is discussed in the following sections, with technical details and insights, to clarify their contributions to improving tile-based SS.

\subsection{Image-based semantic segmentation}

Image-based segmentation extracts image features from the entire scene concurrently and utilizes these global representations to predict the semantic map of the whole image in a truly end-to-end manner.
While tile-based methods dominate current applications, they inherently rely on cropping large-scale RSIs into smaller grids due to GPU memory limitations. 
This artificial physical cropping disrupts genuine geographical topology, causing edge artifacts at tile seams and forcing the truncation of large-scale continuous targets.

To overcome the suboptimal strategies of cropping and fusion, image-based segmentation has recently evolved toward holistic learning paradigms.
Recent breakthroughs, such as REST architecture \cite{chen2025rest}, introduce novel spatial parallel interaction mechanisms. 
These frameworks are specifically designed to handle whole-scene RSIs directly without any cropping operations.
By supporting diverse encoders and decoders in a plug-and-play fashion, such holistic learning architectures enable seamless integration with mainstream segmentation methods and advanced foundation models while bypassing traditional memory bottlenecks.
This paradigm shift resolves the performance degradation associated with local cropping, ensuring that the global contextual dependencies of complex geographical landscapes are fully preserved.

Image-based segmentation represents the ultimate progression in the spatial context modeling hierarchy, shifting the field from local fitting to global cognition.
Although it provides a comprehensive and seamless understanding of massive scenes, this holistic nature places extreme demands on the scalability of computing architectures.
Moving forward, this approach requires deeper exploration into parameter-efficient fine-tuning (PEFT) and the construction of multimodal universal interfaces, enabling these holistic models to efficiently process ultra-large-scale and heterogeneous imagery without degrading back to tile-wise processing.

To comprehensively evaluate these four structural paradigms, Table \ref{GranularityParadigm} summarizes their fundamental characteristics. It explicitly compares the representative algorithms, computational complexity, parameter count, suitable dataset types, supervision requirements, and general generalization capabilities across the pixel, patch, tile, and image hierarchies. This macro-level comparison provides an immediate and intuitive methodological trade-off guide for researchers and engineers.

Furthermore, while this qualitative summary outlines the general theoretical boundaries, comparing absolute quantitative metrics across completely different benchmarks can be inherently misleading if forced into a single dense manuscript table.
To provide rigorous empirical evidence without compromising manuscript readability, we have established a continuously updated open-source performance evaluation database on our GitHub repository at \href{https://github.com/quanweiliu/RSISS}{RSISS}. This repository hosts exhaustive quantitative benchmarks that detail specific accuracy metrics, FLOPs, parameter counts, and processing speeds for representative models discussed throughout this survey.

\section{Feature Extraction Techniques}
\label{FeatureUniExtraction}

Over the past decade, RSISS has undergone significant evolution, establishing a comprehensive spectrum of methodologies that range from pixel-level classification to holistic image-level segmentation. 
Despite the structural differences across these processing granularities, they fundamentally share a core set of underlying feature extraction strategies. 
The dynamic permutations and combinations of these strategies constitute the bedrock of modern remote sensing image semantic segmentation. 
Table \ref{allUniReference} summarizes the primary feature extraction strategies explored in this field. 
In this section, we systematically detail the application of these strategies across pixel, patch, tile, and image data processing paradigms. 
By articulating these shared mechanisms across different hierarchical levels, we aim to inspire future researchers to architect novel feature extraction paradigms and further elevate segmentation performance. 
Because these strategies are predominantly implemented in complex combinations within the existing literature, our subsequent subsections deliberately highlight only the most representative studies to elucidate the core mechanisms of each specific strategy.

\begin{table*}[t]
\caption{\textbf{An overview of feature extraction strategies in remote sensing image semantic segmentation. Based on the surveyed literature, existing feature extraction techniques are further subdivided and organized into several subclasses.}}
\label{allUniReference}
\centering
\small
\renewcommand{\arraystretch}{1.25}
\setlength{\tabcolsep}{4pt}

\begin{tabular}{
p{3.6cm}
p{6.5cm}
p{6.5cm}}
\toprule
\textbf{Feature extraction strategies} &
\textbf{Patch-based feature extraction references} &
\textbf{Tile-based feature extraction references} \\
\midrule

Multi-scale modeling &
Image-level: \cite{tong2020land}, \cite{zhao2016learning}, \cite{gao2020multiscale}, \cite{lee2017going}, \cite{xu2018spectral} \par
Feature-level: \cite{he2017multi}, \cite{pande2022hyperloopnet}, \cite{wang2024ms2canet}
&
Image-level: \cite{guo2018pixel} \par
Feature-level: \cite{zhang2018road}, \cite{li2021multistage}, \cite{zhou2018d}, \cite{kemker2018algorithms}, \cite{li2022a2}, \cite{yi2019semantic} \\
\midrule

Global context extraction &
\cite{yang2021cross}, \cite{hong2022satnet}, \cite{mei2019spectral}, \cite{ma2019double}, \cite{haut2019visual}, \cite{tang2020hyperspectral}, \cite{zhu2020residual}, \cite{sun2019spectral}, \cite{dong2022weighted}, \cite{li2024mambahsi}, \cite{10720896}
&
\cite{ma2023unsupervised}, \cite{wang2021transformer}, \cite{liu2023rethinking}, \cite{wu2025freemix}, \cite{ma2025unified}, \cite{liu2024cm} \\
\midrule

Joint feature learning and loss function &
\cite{xu2018spectral}, \cite{yang2021cross}, \cite{mei2019spectral}, \cite{dong2022weighted}, \cite{zhang2017spectral}, \cite{liu2024universal}, \cite{kang2023self}, \cite{arshad2024light}, \cite{gao2017dual}, \cite{zhong2017spectral}, \cite{zhang2021s3net}, \cite{jia20213}, \cite{roy2023spectral}, \cite{fu2023local}, \cite{song2024joint}
&
\cite{volpi2016dense}, \cite{li2021abcnet}, \cite{wang2022unetformer}, \cite{huang2024decouple}, \cite{li2023spgan}, \cite{sun2022controllable}, \cite{chen2024hi} \\
\midrule

Efficient models &
\cite{arshad2024light}, \cite{zhou2019hyperspectral}, \cite{wang2018fast}, \cite{liu2022fast}, \cite{zhang2019hyperspectral}
&
\cite{li2022a2}, \cite{wang2021transformer}, \cite{liu2024cm}, \cite{li2021abcnet}, \cite{yu2018bisenet} \\
\midrule

Knowledge distillation &
Response-based: \cite{chi2022novel}, \cite{yue2021self} \par
Feature-based: \cite{shi2022explainable}, \cite{zhao2022life}, \cite{li2025hyperkd}
&
Response-based: \cite{dong2023distilling}, \cite{zhou2024mstnet} \par
Feature-based: \cite{zhou2024stonet}, \cite{sun2025handling} \par
Relationship-based: \cite{zhou2023graph}, \cite{zheng2025enhancing} \\
\midrule

Image augmentation &
Transformation-based: \cite{chen2016deep}, \cite{liu2024universal}, \cite{kong2018spectral}, \cite{feng2019divide}, \cite{yu2023hyperspectral}, \cite{liu2020deep}, \cite{chang2024unsupervised}
&
Transformation-based: \cite{li2022global}, \cite{muhtar2022index}, \cite{10309935}, \cite{wang2023fine} \par
Generation-based: \cite{benjdira2019unsupervised}, \cite{tasar2020colormapgan}, \cite{ji2020generative}, \cite{cai2022iterdanet} \\
\midrule

Transfer learning and fine-tuning  &
\cite{yang2017learning}, \cite{huang2018urban}, \cite{he2019heterogeneous}, \cite{zhong2020cross} 
&
\cite{zhang2024rs}, \cite{lu2024multi}, \cite{wang2023samrs}, \cite{wu2024compositional}, \cite{chen2025multi}, \cite{yang2025high}, \cite{ma2025unified} \\
\midrule

Meta-learning &
\cite{liu2018deep}, \cite{gao2022unsupervised}, \cite{li2022deep}, \cite{zhang2022cross}, \cite{xu2023graph}, \cite{zeng2023multistage}, \cite{li2024scformer}, \cite{feng2024cross}
&
\cite{jiang2022fewshotseg}, \cite{wang2022dmmlnet}, \cite{lang2023r2net}, \cite{cao2024frinet}, \cite{puthumanaillam2023texture}, \cite{li2024mgcl}, \cite{chen2024tafd}, \cite{li2024segland} \\
\midrule

Domain adaptation &
Feature-level: \cite{xu2023graph}, \cite{zhang2021topological}, \cite{zhang2022graph}, \cite{zhao2022cross}, \cite{huang2022two}, \cite{long2015learning}, \cite{liu2020class}, \cite{gao2025pseudo}, \cite{damodaran2018deepjdot}, \cite{he2025hyper}, \cite{ye2024building}, \cite{xin2024feature}, \cite{feng2024s4dl} \par
Output-level: \cite{tong2020land}
&
Input-level: \cite{benjdira2019unsupervised}, \cite{tasar2020colormapgan}, \cite{ji2020generative}, \cite{cai2022iterdanet}, \cite{tasar2020standardgan}, \cite{wittich2021appearance}, \cite{peng2021full} \par
Feature-level: \cite{ma2023unsupervised}, \cite{wang2023fine}, \cite{lu2021cross}, \cite{wu2022deep}, \cite{zhang2020unsupervised} \par
Output-level: \cite{zheng2021entropy}, \cite{chen2022unsupervised}, \cite{li2022unsupervised}, \cite{wang2022cross}, \cite{hoyer2022daformer}, \cite{zhu2023unsupervised}, \cite{ma2023domain}, \cite{zhang2021curriculum} \\
\midrule

Domain generalization &
Data manipulation: \cite{zhang2023single}, \cite{zhao2023locally}, \cite{dong2024spectral}, \cite{cai2025dynamicstyle} \par
Domain-invariant feature representation: \cite{zhang2023ldgnet}, \cite{li2025scpnet}, \cite{qin2024fdgnet}, \cite{gao2025isdgs}, \cite{han2024mscdg} \par
General learning strategy: \cite{wang2024hypersigma}, \cite{zhang2025spectralx}
&
Data manipulation: \cite{iizuka2023frequency}, \cite{liang2024ccdr} \par
General learning strategy: \cite{wu2025freemix}, \cite{wang2025stylemap}, \cite{luo2025geosabasa}, \cite{yaghmour2025sensoragnostic}, \cite{zhao2025crlmdg}, \cite{gong2024crossearth} \\
\midrule

Self-supervised learning &
Self-predictive generative learning: \cite{wang2024hypersigma}, \cite{mou2017unsupervised}, \cite{mei2019unsupervised}, \cite{ibanez2022masked}, \cite{scheibenreif2023masked}, \cite{DBLP00953} \par
Discriminative contrastive learning: \cite{yu2023hyperspectral}, \cite{liu2020deep}, \cite{chang2024unsupervised}, \cite{DBLP00953}, \cite{app11188670}, \cite{hou2021hyperspectral}, \cite{guan2022cross}
&
Self-predictive generative learning: \cite{li2021semantic}, \cite{sun2022ringmo}, \cite{cong2022satmae}, \cite{cai2024consistency}, \cite{liu2023rethinking}, \cite{cao2023transformer} \par
Discriminative contrastive learning: \cite{li2022global}, \cite{muhtar2022index}, \cite{li2021semantic}, \cite{cao2023transformer}, \cite{rs13163275}, \cite{dong2023spatial} \\
\midrule

Semi-supervised learning &
Graph-based: \cite{dong2022weighted}, \cite{liu2020cnn} \par
Pseudo-labelling/Self-training: \cite{wu2017semi}, \cite{seydgar2022semisupervised} \par
Joint optimization: \cite{liu2024universal}, \cite{liu2017semi}, \cite{hang2020adversarial}, \cite{huang2023spectral}
&
Pseudo-labelling/Self-training: \cite{10329942}, \cite{wang2022semi}, \cite{lu2022simple} \par
Consistency regularization: \cite{chen2023semi}, \cite{huang2024decouple}, \cite{10309935}, \cite{lu2022simple}, \cite{li2021semisupervised}, \cite{jin2024dynamic}, \cite{wang2020semi}, \cite{wang2021ranpaste}, \cite{lv2024advancing}, \cite{chen2024category}, \cite{chen2022semi}, \cite{guo2023semisupervised}, \cite{kang2021picoco}, \cite{rs14040879} \\
\midrule

Weakly-supervised learning &
\cite{yang2023iter}
&
\cite{fu2018wsf}, \cite{CAO2022157}, \cite{fang2022improved}, \cite{li2021effectiveness}, \cite{chen2020spmf}, \cite{iqbal2020weakly}, \cite{lu2022nfanet}, \cite{9372390}, \cite{LI2022244} \\
\bottomrule
\end{tabular}
\end{table*}

\begin{figure}[t]
\centering
\subfloat[\textbf{Image-level multiscale feature extraction.}]{\includegraphics[width=1.5 in]{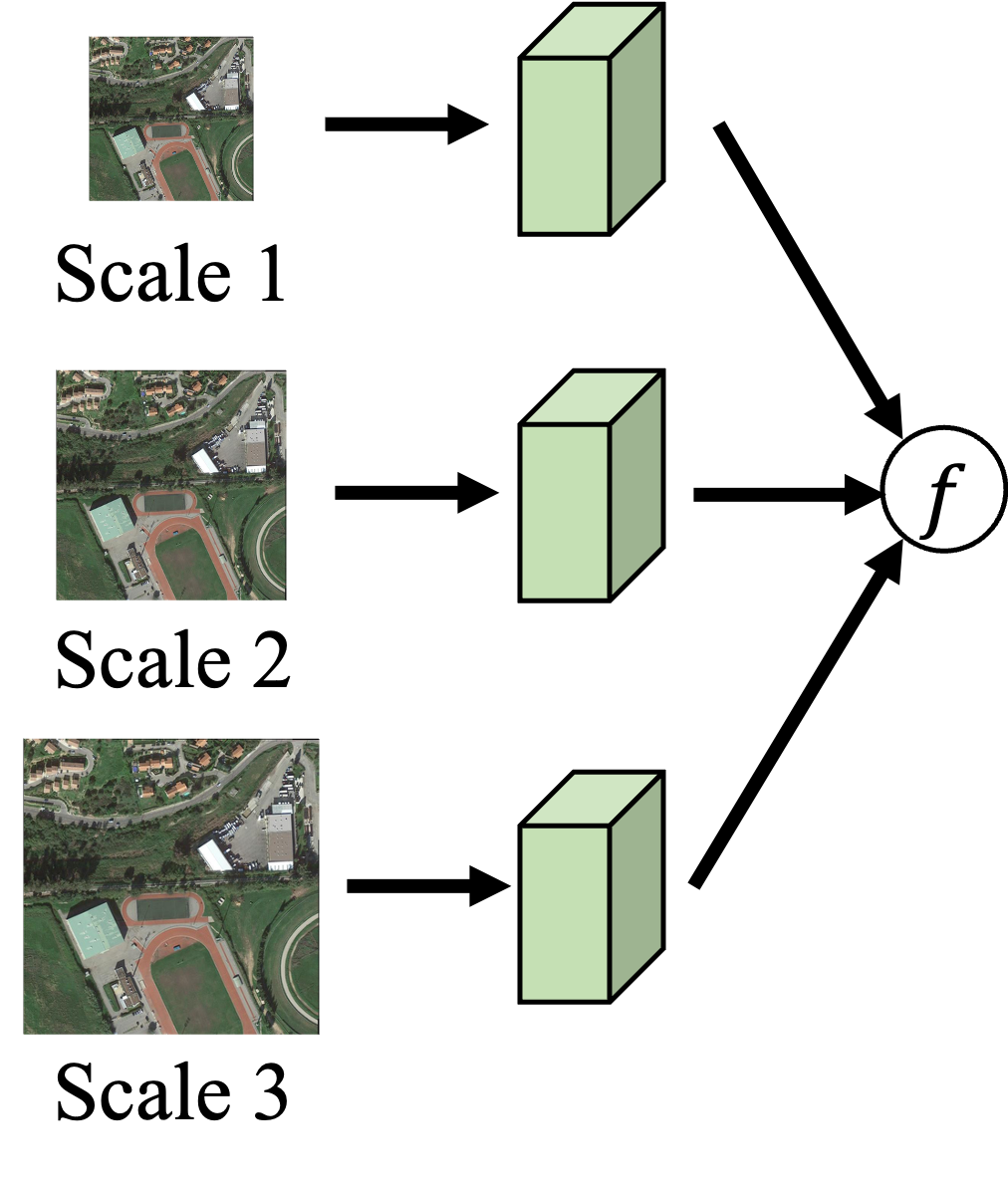}}
\hspace{6mm}
\subfloat[\textbf{Feature-level multiscale feature extraction.}]{\includegraphics[width=1.5 in]{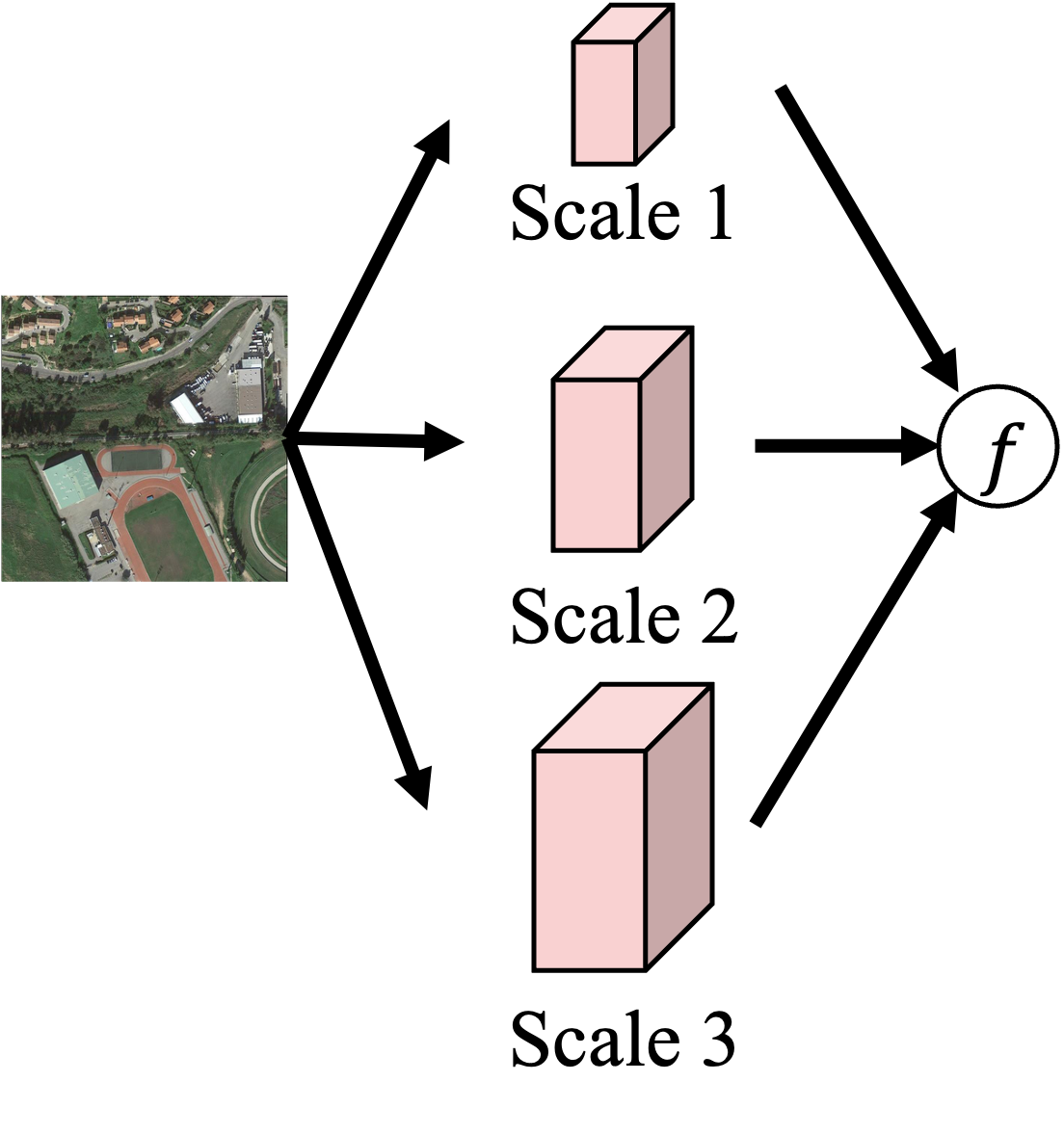}}
\caption{\textbf{Two representative multiscale feature extraction paradigms. (a) Image-level multiscale feature extraction, where the input image is rescaled into multiple resolutions and processed independently before feature aggregation. (b) Feature-level multiscale feature extraction, where multiscale representations are generated within the network from a single-resolution input and subsequently fused.}}
\label{multiscale}
\end{figure}

\subsection{Multi-scale modeling}

Object scales vary dramatically in RSIs, and constructing robust multi-scale representations is particularly important.
Various DL models strive to extract valuable information for target perception from limited receptive fields.
CNNs extract local spatial features, and only through residual networks can the receptive field be gradually expanded to learn more complex spatial patterns \cite{slavkovikj2015hyperspectral, duan2017sar}.
The capsule network (CapsNet) \cite{wang2020novel} captures positional and directional dependencies between features, thereby providing more detailed feature representations.
Additionally, GNNs \cite{dong2022weighted} can capture spatial dependencies from non-Euclidean data structures, such as graphs, offering an efficient alternative for spatial feature extraction.

However, clear scale differences among various objects in RSIs. A more effective strategy for enhancing feature extraction is to capture spatial dependencies from multi-scale contexts, which enables the model to gather information at various levels, which is particularly important when handling objects of different scales.
Common techniques for achieving multi-scale information include pyramid-based feature extraction and multi-branch feature fusion methods, which can be reduced to the image-level \cite{tong2020land, zhao2016learning, gao2020multiscale, lee2017going, xu2018spectral, guo2018pixel} and feature-level \cite{he2017multi, pande2022hyperloopnet, wang2024ms2canet, zhang2018road, li2021multistage, zhou2018d, kemker2018algorithms, li2022a2, yi2019semantic} methods, as depicted in Figure \ref{multiscale}.

For patch-wise classification, the relatively small input patch size has led many studies to adopt multi-scale image cropping before feeding data into the model.
For instance, Zhao and Du \cite{zhao2016learning} introduce a multiscale CNN with image pyramids and spectral–spatial feature fusion via logistic regression, improving urban classification performance.
Gao et al. \cite{gao2020multiscale} propose a lightweight multiscale residual network using mixed depthwise separable convolutions and high-level shortcuts to extract spectral–spatial features.

Meanwhile, multi-scale feature fusion is widely used in tile-wise SS.
UNet-like models merge features from different depths, which can also be regarded as a form of multi-scale contextual representations \cite{zhang2018road, li2021multistage, li2022a2, yi2019semantic}.
For example, DeepResUnet \cite{yi2019semantic} follows U-Net’s encoder–decoder, extracting multiscale features via cascaded downsampling and skip-concatenation, while residual blocks deepen each scale for robust building segmentation.
MAResU-Net \cite{li2021multistage} replaces plain skip connections in U-Net with multi-stage linear-attention blocks that refine multiscale encoder–decoder features and inject global context during stage-wise fusion in decoding.

Many RSISS methods are directly adapted from computer vision architectures to obtain multi-scale features.
For example, D-LinkNet, based on the LinkNet architecture \cite{chaurasia2017linknet}, incorporates dilated convolutions to improve road extraction performance \cite{zhou2018d}.
Kemker et al. \cite{kemker2018algorithms} adapted the Sharpmask and RefineNet models to process MSI data.
The atrous spatial pyramid pooling (ASPP) has also been adopted to capture RSI features at multiple spatial scales \cite{guo2018pixel}.

In summary, multi-scale modeling has emerged as a simple yet effective feature extraction strategy and is now regarded as a fundamental design principle in modern RSISS architectures. 
Many networks implicitly incorporate multi-scale feature extraction in their design, even when it is not explicitly highlighted. 
However, multi-scale designs often introduce substantial computational overhead. Consequently, developing lightweight multi-scale feature extraction mechanisms while maintaining strong segmentation performance remains a key direction for future research.

\subsection{Global context extraction}
 
For humans, objects are categorized based on their shape, texture, and other macroscopic features, where contextual information plays a crucial role in determining object categories.
In RSISS models, global context is typically captured through (i) pooling-based modules and (ii) attention mechanisms.

\begin{figure}[!t]
    \centering
        \includegraphics[width=3 in]{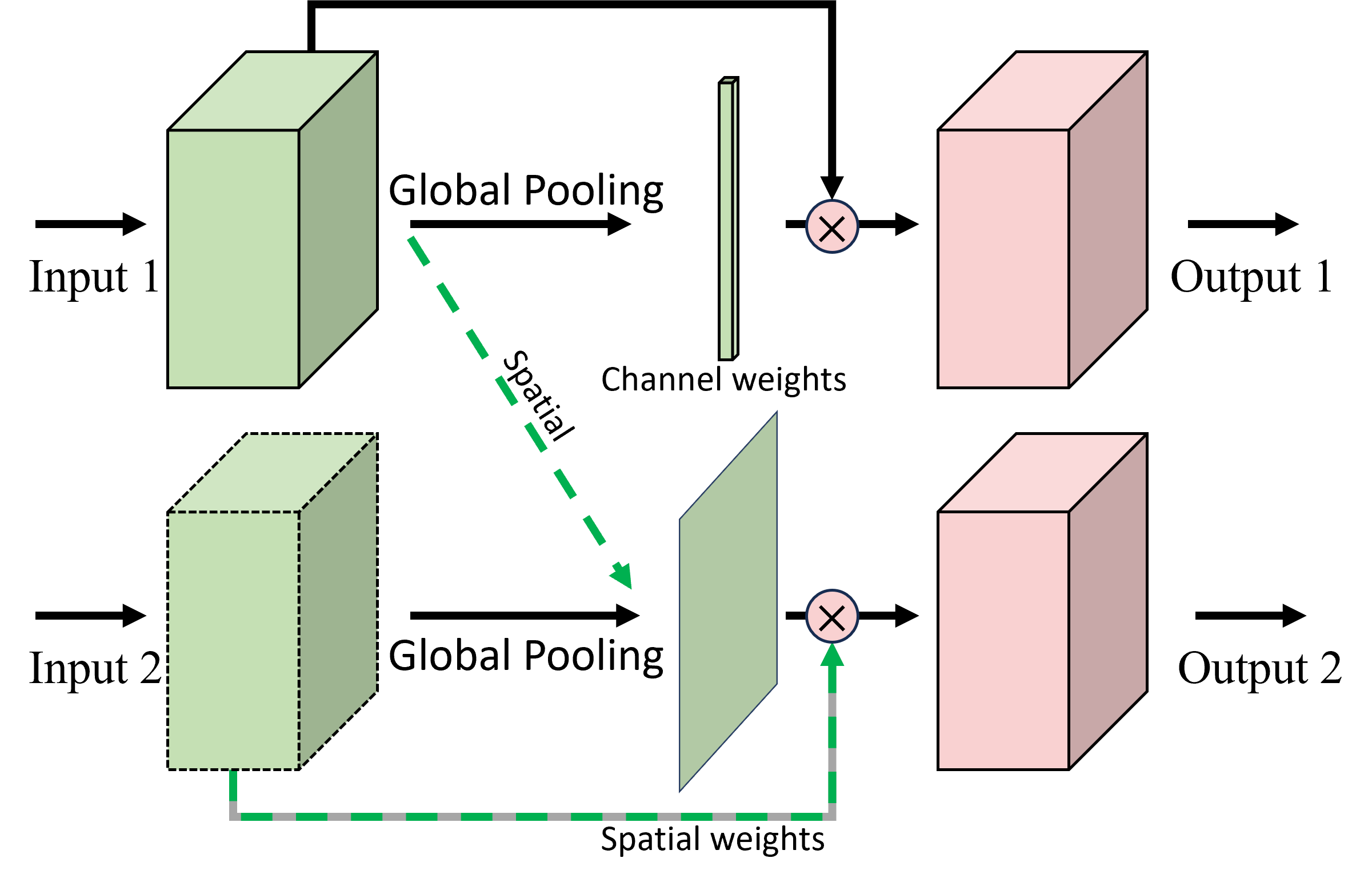}
        \caption{\textbf{Illustration of attention mechanisms in feature enhancement. The top branch shows channel attention, where global pooling generates channel-wise weights to recalibrate feature responses. The bottom branch shows spatial attention, where spatial importance maps are computed to emphasize informative regions. The green dashed lines indicate cross-attention interactions between channel and spatial representations.}}
    \label{implicitAttention}
\end{figure}

While global pooling offers a simple and efficient solution, it often leads to the loss of small targets due to aggressive downsampling.
In contrast, attention mechanisms provide a more flexible and effective means to capture long-range dependencies by assigning adaptive weights across the image space.
It enables the model to focus selectively on relevant parts of the input, thereby improving the capture of dependencies and contextual relationships.
The most widely used attention strategies include channel attention, spatial attention, and self-attention mechanisms \cite{yang2021cross, hong2022satnet, wang2021transformer, ma2023unsupervised}.
Figure~\ref{implicitAttention} presents an overview of the channel and spatial attention mechanisms, while Figure~\ref{explicitAttention} illustrates the self-attention mechanism framework.

For high-dimensional data, increasing studies have been given to integrate both spectral and spatial attention for feature extraction \cite{mei2019spectral, ma2019double, haut2019visual, zhu2020residual, sun2019spectral}.
Mei et al. \cite{mei2019spectral} employed RNNs with attention to capture intrinsic spectral correlations.
Ma et al. \cite{ma2019double} proposed a dual-branch structure to extract spectral and spatial features separately, applying distinct attention mechanisms in each branch.
This parallel configuration allows for the independent optimization of complementary feature sets prior to fusion \cite{tang2020hyperspectral}.
In \cite{zhu2020residual}, spatial and spectral attention modules are arranged in a cascaded structure within a residual block framework, a design well-suited for refining features progressively across network layers \cite{sun2019spectral}.

In high-resolution data, global feature extraction is crucial for capturing long-range spatial dependencies that CNNs alone often miss. 
Wang et al. \cite{wang2021transformer} introduced a bilateral design, where a CNN branch preserves fine-grained local details and a Transformer branch models long-range interactions to handle large appearance variations in very fine-resolution urban scenes. In a related direction, MBATA-GAN \cite{ma2023unsupervised} adopted mutually boosted attention to strengthen cross-domain global interactions over high-level features. 
More recently, RSISS increasingly relies on Transformer-centric pipelines and foundation models to enhance global context modeling. 
GLOTS \cite{liu2023rethinking} leverages a pretrained ViT encoder and a global–local attention decoder to inject global cues into multi-scale representations, while FreeMix \cite{wu2025freemix} builds on a pretrained foundation backbone and a Mask2Former-style Transformer decoder to propagate global semantics to dense masks under domain shifts. 
MFNet \cite{ma2025unified} further demonstrates that fine-tuning SAM-style ViT encoders and coupling them with explicit global channel aggregation can provide strong global priors for high-resolution multimodal segmentation.

In addition, the Mamba network has been adopted in various computer vision tasks for its efficiency in integrating both global and local contextual information \cite{li2024mambahsi, 10720896}.
It achieves this without incurring the high computational cost typically associated with traditional self-attention mechanisms \cite{liu2024cm}.
Mamba belongs to the family of selective state space models, which model long-range dependencies via recurrent state updates and input-dependent filters. 
In principle, each token’s state summarizes information from all previous tokens, leading to a global receptive field in the sequence dimension.

\subsection{Joint feature learning and loss function}

By designing specific neural network architectures and incorporating customized loss functions, networks can be optimized to extract multiple types of information in a targeted manner, such as spatial, spectral, local, and global information.
Joint feature learning has been shown to significantly enhance segmentation performance.

At the data level, HSIs provide both spatial and rich spectral features \cite{kang2023self, arshad2024light, zhong2017spectral}, while polarimetric SAR imagery enables decomposition into polarization and spatial components \cite{gao2017dual}.
Zhong et al. \cite{zhong2017spectral} incorporate CNN structures to extract spectral and spatial features from HSIs.
Gao et al. \cite{gao2017dual} constructed a dual-branch deep CNN where polarization features were extracted from a six-channel real matrix and spatial features from a Pauli RGB image.
More generally, heterogeneous network architectures are often employed to extract joint features from separate pathways \cite{zhang2021s3net, yang2021cross, mei2019spectral}.
For example, Zhang et al. \cite{zhang2017spectral} used a 1-D CNN to extract spectral features and a 2-D CNN to extract spatial features.
SSUN \cite{xu2018spectral} utilised LSTM networks for spectral feature extraction and a multi-scale CNN for spatial features.

At the feature level, CNNs are effective in capturing local patterns, while attention mechanisms extract global context \cite{fu2023local}.
Jia et al. \cite{jia20213}  proposed 3-D Gabor-modulated kernels to characterize the internal spatial–spectral structure of HSI data from various perspectives.
Roy et al.  \cite{roy2023spectral} proposed a morphFormer that implements a learnable spectral and spatial morphological network. The morphological convolution operations enhanced the interaction between shape and structural information.
In addition, semantic labels convey category-level information, and edge labels provide spatial boundary details \cite{song2024joint}.
The integration of these features helps to delineate semantic regions more precisely, improving segmentation accuracy.
Song et al. \cite{song2024joint} applied a Laplacian filter to extract edge features as an explicit supervisory signal, incorporating a segmentation head and edge decoder into the network to jointly learn semantic and boundary features and enhance generalisation.

The loss function measures the discrepancy between the model’s predicted output and the ground truth. Modifying the loss function can guide the model's optimization, enabling it to learn specific features that aid segmentation.
Commonly used loss functions in SS include cross-entropy loss, dice loss, focal loss, and infoNCE loss \cite{volpi2016dense, li2021abcnet, wang2022unetformer}, and their weighted summation \cite{dong2022weighted, liu2024universal}.
Specialized loss functions are integral to constructing semi-supervised learning (SeL), self-supervised learning (SSL), and multi-task learning frameworks, enabling the model to extract meaningful information from data.
In these cases, losses such as adversarial loss, cycle consistency loss, perceptual loss, local consistency loss, and global diversity metric loss are commonly employed \cite{huang2024decouple, li2023spgan}.
Additionally, edge loss is frequently used as an auxiliary objective to enhance boundary delineation \cite{sun2022controllable, chen2024hi}.
This strategy aids in detecting small objects and distinguishes between classes with similar shapes.

\subsection{Efficient models}
Due to differences in learning styles, such as recurrent, transductive, and inductive learning \cite{Mao_2021_ICCV}, DL architectures vary significantly in time or space requirements.
Technically, RNNs, which follow a recurrent learning style, are generally slow during training due to their loop structure but offer fast inference once trained \cite{zhou2019hyperspectral}.
SeL GNNs operate under a transductive learning paradigm.
Their shallow architecture enables fast training and inference, though they require considerable memory for processing \cite{dong2022weighted}.
An efficient GNN such as FDGC \cite{liu2022fast} used inductive style training, achieving a balance between speed and efficiency, but lost the ability to extract global non-Euclidean distance relationships.
CNNs, Transformers, and Mamba models are usually used as inductive models. 
Although these models require more time per training unit, they can achieve convergence using only tens or hundreds of samples per class in patchwise classification tasks, often resulting in overall faster training \cite{wang2018fast}.

In pixel- and patch-level classification, overlapping patches from neighbouring pixels introduce significant redundancy in computation, thereby inevitably extending test time \cite{arshad2024light, zhang2019hyperspectral, sandler2018mobilenetv2}.
In tilewise segmentation, the scale invariance of inputs and outputs addresses the issue of redundant computations during the inference phase \cite{long2015fully}.
However, recent advanced DL models for segmentation frequently involve complex architectures and require large training datasets, which result in substantial computational demands.
As a result, computational efficiency in tile-based segmentation primarily concerns the number of parameters and training speed.

To construct efficient models, lightweight modules such as depthwise convolution, pointwise convolution \cite{sandler2018mobilenetv2}, linear attention \cite{li2022a2}, bilateral segmentation networks \cite{yu2018bisenet}, and Mamba modules \cite{liu2024cm} are widely adopted.
These components help reduce computational cost while having only a minor impact on segmentation accuracy or none at all.
For example, linear attention mechanisms have been integrated into lightweight bilateral contextual networks to improve computational efficiency \cite{li2022a2, wang2021transformer, li2021abcnet}.
CM-UNet \cite{liu2024cm} combines a CNN-based encoder with a Mamba-based decoder to efficiently extract and integrate both local and global features for RSISS.
This design improves segmentation performance while maintaining a low computational footprint.

\subsection{Knowledge distillation (KD)}

Among the efficient building blocks for deep models, KD has been proposed as an effective solution for transferring and refining the knowledge of large models to small models \cite{hinton2015distilling}. 
KD typically enables lightweight student networks to mimic the behaviour of larger, well-trained teacher models, achieving competitive or even superior performance with lower complexity. 
These methods offer practical pathways to reduce model complexity while maintaining accuracy and lay the foundation for more robust models in scenarios with limited annotations or dynamic domain shifts.
KD has evolved from basic logit matching to adaptive, interpretable, and lifelong knowledge transfer, including response-based \cite{chi2022novel, yue2021self, dong2023distilling, zhou2024mstnet}, feature-based \cite{shi2022explainable, zhao2022life, li2025hyperkd, zhou2024stonet, sun2025handling}, and relationship-based \cite{zhou2023graph, zheng2025enhancing} approaches based on the different knowledge categories of teacher models.

\noindent\textbf{Response-based KD.}
One major line of work focuses on response-based KD, in which soft output logits from the teacher guide the student \cite{chi2022novel, yue2021self}. 
For example, Chi et al. \cite{chi2022novel} proposed SSKD, a self-supervised framework that generates soft labels for unlabeled HSI patches using spectral–spatial similarity, effectively leveraging large-scale unlabeled data for supervision. 
In DSCT, Dong et al. \cite{dong2023distilling} proposed a hybrid CNN–Transformer distillation framework with a novel target–nontarget KD strategy that explicitly guides decision boundary refinement. 
Likewise, MSTNet-KD \cite{zhou2024mstnet} employs multilevel output alignment to bridge deep decoder layers between student and teacher networks.

\noindent\textbf{Feature-based KD.}
Feature-based KD captures structural representations from internal layers of the teacher model. 
Shi et al. \cite{shi2022explainable} proposed an explainable scale distillation network that transfers multi-scale information from a complex teacher to a single-scale student. The distilled knowledge preserves both the discriminative power and interpretability of multi-scale representations while reducing computational cost. 
Zhao et al. \cite{zhao2022life} further extended KD to lifelong learning by proposing a continual spectral–spatial feature distillation strategy that maintains knowledge across sequential HSI tasks without catastrophic forgetting. 
Complementary to this, Li et al. \cite{li2025hyperkd} developed HyperKD, which combines exemplar replay with cross-spectral–spatial KD, allowing the student model to inherit not only output predictions but also spectral and spatial distributions from previous tasks.
In STONet-S, Zhou et al. \cite{zhou2024stonet} introduced frequency-aware KD using discrete cosine transforms to decompose and transfer high- and low-frequency components, preserving both edge and semantic information. 
Additionally, Sun et al. \cite{sun2025handling} addressed the robustness of KD under weak supervision. The proposed BAKD framework combines boundary-aware and uncertainty-weighted distillation to reduce the impact of noisy annotations, especially near semantic edges.

\noindent\textbf{Relationship-based KD.}
Relationship-based KD has also gained traction. Graph-aware methods such as GAGNet-S transfer topological context by encoding cross-scale and inter-pixel dependencies \cite{zhou2023graph}. 
Meanwhile, AKD \cite{zheng2025enhancing} strategies in transformer-based dual-path networks enhance student learning by minimizing the angle between teacher–student feature vectors across channels. 
Overall, KD research in RSISS demonstrates a clear trend toward structure-aware, frequency-guided, and uncertainty-adaptive strategies to enhance generalization while reducing model complexity.

\subsection{Image augmentation}
The primary function of image augmentation is to artificially increase the size and diversity of training datasets, thereby improving model generalization and robustness. 
Since collecting and annotating large-scale datasets is both costly and time-consuming, image augmentation generates additional samples by perturbing existing images while preserving their semantic labels. 
This helps segmentation models cope with variations in environment, illumination, sensor characteristics, and viewpoint, without sacrificing pixel-level detail.

\noindent\textbf{Transformation-based image augmentation.}
From a transformation-based perspective, augmentation applies predefined geometric or radiometric operations to existing images. 
Typical choices include random flipping, cropping, rotation, jittering, and random erasing, which introduce local variations but keep the global scene structure unchanged \cite{chen2016deep, kong2018spectral, feng2019divide}. 
For modality-specific data such as HSI and LiDAR, specialized operations have been designed, for example, adding band-wise noise, simulating virtual samples via random spectral scaling \cite{chen2016deep}, or selecting unlabeled hyperpixels around labelled seeds to densify training sets \cite{feng2019divide}.
In self-supervised learning, such transformations are further organized into pairs or groups to construct positive and negative views for contrastive objectives, making augmentation an integral part of representation learning rather than a mere pre-processing step \cite{liu2024universal, yu2023hyperspectral, liu2020deep, chang2024unsupervised, zhou2024mstnet}.
In consistency regularization, SeL and KD, pairwise images generated through data augmentation can also guide models to learn invariant feature representations and complementary knowledge \cite{wang2023fine, 10309935, dong2023distilling}.
For example, ECAE \cite{10309935} uses color jitter and Gaussian blur perturbations to enforce consistency when learning from unlabeled data, improving pseudo-labels and boundaries. 
DSCT \cite{dong2023distilling} trains distilled CNN/transformer students with random resizing, cropping, flipping, and photometric distortions, boosting robustness and generalization across datasets.

\noindent\textbf{Generation-based image augmentation.}
Beyond these handcrafted transformations, generation-based augmentation leverages generative models to synthesize new training images or re-style existing ones. 
In RSISS, this is often coupled with unsupervised DA: the generator maps labeled source images into target-style images while preserving semantic structure, effectively creating additional supervision for the target domain \cite{benjdira2019unsupervised, tasar2020colormapgan, ji2020generative, cai2022iterdanet}. 
Benjdira et al. \cite{benjdira2019unsupervised} demonstrated this idea by training a CycleGAN-based image-to-image translation network. 
Tasar et al. \cite{tasar2020colormapgan} further simplified the generative component with ColorMapGAN, which learns a global color mapping between source and target images. 
This design produces “fake” training images that are semantically identical to the originals yet spectrally aligned with the test domain. 
Ji et al. \cite{ji2020generative} generalized this line of work with a full-space DA network, in which adversarial learning is performed jointly across the image, feature, and output spaces.

Overall, transformation-based and generation-based augmentation are complementary: the former enriches local appearance and geometric diversity around a fixed data distribution, whereas the latter synthesises target-style or domain-bridging samples that explicitly reshape the training distribution. 
For RSISS, especially in cross-city or cross-sensor scenarios, GAN-driven, generation-based augmentation is a powerful tool for narrowing domain gaps while reusing existing pixel-wise annotations, thereby enhancing the practicality of deep segmentation models in real-world deployments.

\subsection{Transfer learning and fine-tuning}

Transfer learning refers to the reuse or adaptation of a pre-trained model for new tasks. 
Rather than training a model from scratch, transfer learning enables the application of learned knowledge, significantly reducing training time and resource demands, particularly when limited data are available for the target task \cite{yang2017learning, huang2018urban, he2019heterogeneous, zhong2020cross}. 
For instance, Zhong et al. \cite{zhong2020cross} integrated transfer learning with DA to reduce distributional discrepancies across scenes and sensor types, proposing a cross-scene transfer learning approach that performs well even with a small number of labelled samples.
One issue in transfer learning is the domain shift between the original data and fine-tuning data. To overcome this, He et al. \cite{he2019heterogeneous} introduce a mapping layer that projects the HSI to 3-D features and then uses a pretrained CNN to initialize the network.

The rapid emergence of large-scale VFMs, such as ViT-based backbones and the Segment Anything Model (SAM) \cite{kirillov2023segment}, has fundamentally changed how transfer learning is applied in RSISS. 
Directly fine-tuning all parameters of these models on relatively small RS datasets is often inefficient and prone to overfitting, due to the large number of parameters, high memory cost, and substantial training time. 
As a result, PEFT has become increasingly favored over conventional full-model fine-tuning. 
Instead of updating the entire backbone, PEFT freezes most pre-trained weights and inserts lightweight, task-specific modules, such as adapters, low-rank adapter branches, or multi-level feature adaptation blocks, thereby improving both accuracy and efficiency \cite{wang2023samrs, ma2025unified, zhang2024rs, lu2024multi, wu2024compositional, chen2025multi, yang2025high}.

Recent SAM-based RSISS methods exemplify this shift. 
Multi-LoRA \cite{lu2024multi} fine-tuned SAM frameworks for urban man-made object extraction, which introduces multiple LoRA modules into the SAM encoder and decoder, enabling accurate road and building extraction while updating only a small fraction of parameters. 
Other works propose multi-level feature adaptation of SAM for high-resolution land use/land cover classification \cite{yang2025high}, or design unified multimodal fine-tuning frameworks that plug multimodal fusion heads on top of frozen VFMs to jointly exploit optical, SAR, and LiDAR data \cite{ma2025unified}.

Overall, the role of transfer learning in RSISS is evolving from classical “pretrain on natural images + full fine-tuning” toward adapter- and PEFT-driven fine-tuning of foundation models. 
Traditional transfer learning remains effective for moderate-sized CNNs, but for large VFMs, it is increasingly replaced or complemented by PEFT strategies that offer a better trade-off between accuracy, efficiency, and generalization across diverse RS scenes and modalities.

\begin{figure}[!t]
    \centering
        \includegraphics[width=3 in]{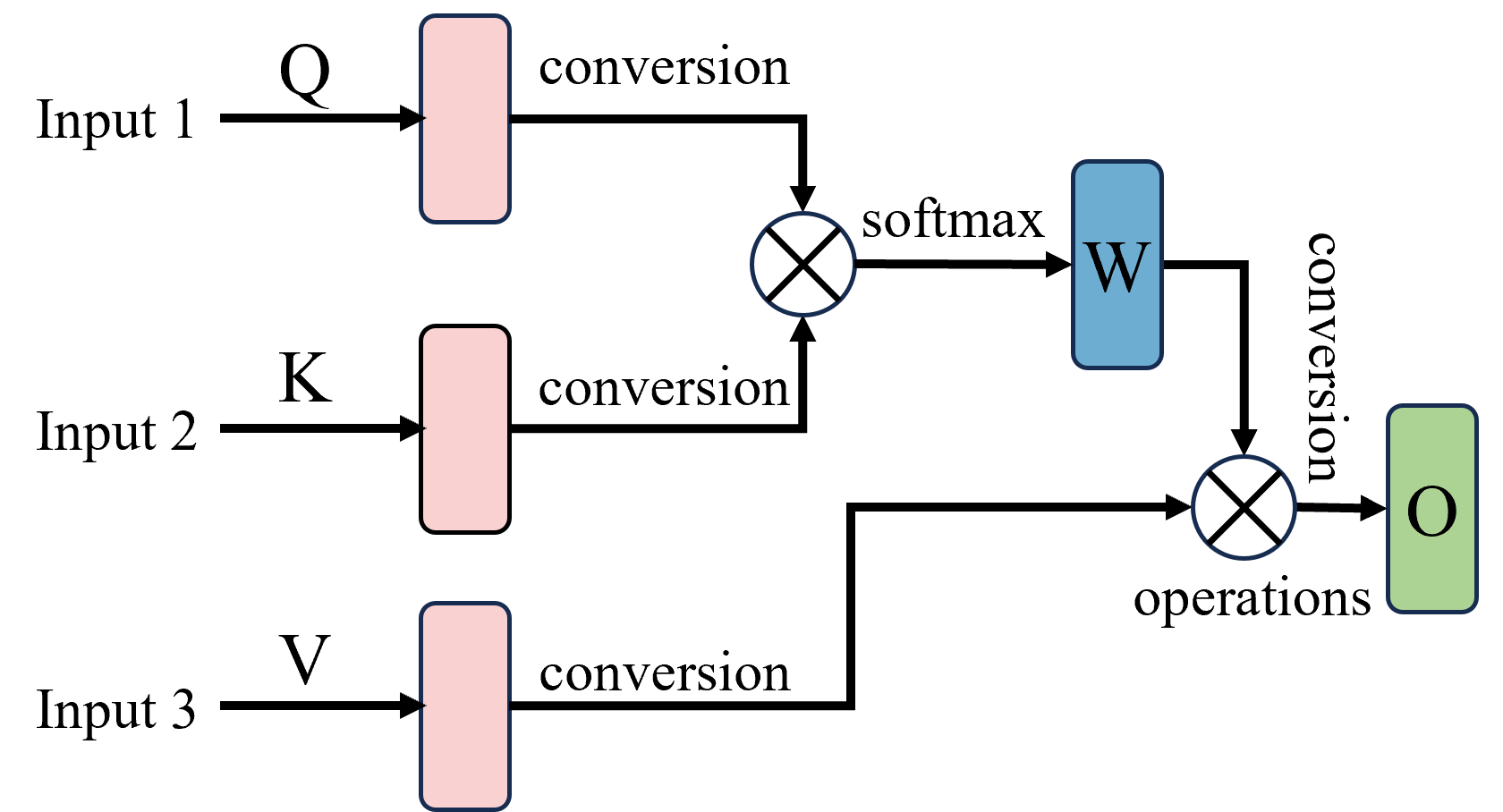}
        \caption{\textbf{Self-attention and cross-attention mechanisms based on the query–key–value (Q–K–V) framework. Query (Q), key (K), and value (V) features are first projected through linear transformations. Attention weights are computed via similarity between Q and K followed by a softmax operation, and are then applied to V to produce the output representation. In self-attention, Q, K, and V originate from the same feature source, whereas in cross-attention, they are derived from different feature sources or modalities.}}
    \label{explicitAttention}
\end{figure}

\subsection{Meta-learning}
Meta-learning enables models to adapt quickly to new tasks by training on a variety of related tasks.
The objective is to learn a generalizable strategy that enables the model to perform well on new tasks with minimal data and few training iterations \cite{luo2022meta}.

In HSI classification, meta-learning has become a central tool for tackling small-sample and cross-domain settings.
Early works employed residual 3-D CNNs and episodic training to learn a discriminative metric space in which query pixels are classified by nearest neighbors to class prototypes \cite{liu2018deep}.
To further reduce dependence on annotated sources, unsupervised meta-learning with multiview constraints has been proposed, which constructs tasks from unlabeled HSIs and trains relation networks to learn transferable metrics \cite{gao2022unsupervised}.
Building on this foundation, a series of cross-domain few-shot HSI classification methods explicitly integrate DA into the meta-learning pipeline, including deep cross-domain metric learning with adversarial alignment \cite{li2022deep}, contrastive few-shot learning that enhances inter-class separation in highly confounded spectral spaces \cite{zhang2022cross}, graph-based DA that leverages graph convolutions over spectral–spatial prototypes \cite{xu2023graph}, and multistage relation networks with dual metrics to improve class representation under extremely scarce labels \cite{zeng2023multistage}.
More recent transformer-based and distillation-based approaches further push this line of research by injecting spectral coordinate priors into asymmetric encoder–decoder architectures and by decoupling meta-knowledge distillation across domains \cite{li2024scformer, feng2024cross}, yielding strong cross-domain few-shot HSI classification performance.

In parallel, meta-learning has been extended from patch-level classification to tile-level SS in RSIs, particularly in few-shot settings.
A first step was to formulate RSISS as a prototype-based few-shot segmentation approach, in which deep metric learning is used to extract pixel embeddings, construct class prototypes from support masks, and segment query images via nearest-prototype matching in the embedding space \cite{jiang2022fewshotseg}.
Subsequent works refine this metric-based meta-learning paradigm to better accommodate the unique characteristics of high-resolution aerial and satellite images \cite{wang2022dmmlnet, lang2023r2net, cao2024frinet}.



Overall, existing meta-learning studies in RS are better viewed as several complementary research streams rather than a single linear trajectory.
One stream focuses on task-level episodic meta-learning for HSI, where models are trained to rapidly adapt to new spectral classes or scenes under scarce labels.
A second stream emphasizes metric- and optimization-based few-shot learning with explicit cross-domain robustness, in which adversarial alignment, graph modeling, transformers, and knowledge distillation are integrated into the meta-learning pipeline to handle shifts across sensors, regions, and acquisition conditions.
A third stream extends these ideas from classification to segmentation, yielding few-shot segmentation frameworks tailored to high-resolution RGB and multispectral RSIs.
Despite the diversity of streams, these works share a common core meta-learning principle: by simulating many support–query episodes during training, the model acquires an inductive bias that enables rapid adaptation to new categories, domains, and geographic regions from only a handful of annotated examples.

\subsection{Domain adaptation (DA)}
\begin{figure}[t]
\centering
\includegraphics[width=3.5 in]{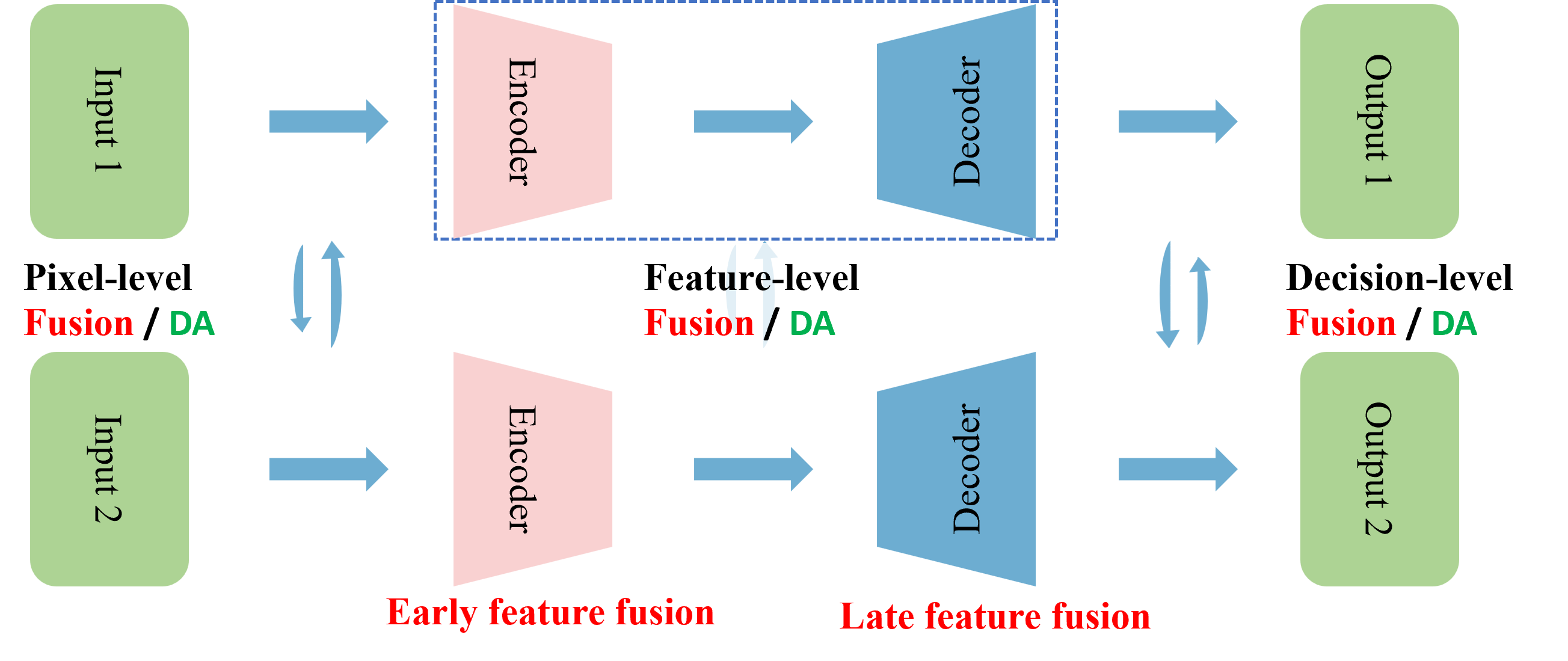}
\caption{\textbf{Unified remote sensing image semantic segmentation framework incorporating fusion and domain adaptation (DA) at multiple levels. Fusion can be performed at the pixel level (input fusion), feature level (early or late feature fusion within encoder–decoder architectures), or decision level (output fusion). DA can be applied at the pixel, feature, or decision stages to enhance cross-domain robustness.}}
\label{fusionFramework}
\end{figure}

Due to sensor nonlinearities, seasonal variation, or weather differences, it is common for training and test sets to be collected under different conditions.  
The discrepancy between the data distribution of the source domain (SD) and the target domain (TD) is known as a distribution shift or domain gap.
DA, typically in the form of unsupervised domain adaptation (UDA), aims to leverage data from both the SD and TD during training to learn shared knowledge across the input, feature, and output spaces, reducing the domain gap and improving model performance on the TD \cite{benjdira2019unsupervised, tasar2020colormapgan, zhang2021topological, zhang2022graph}.
As illustrated in Figure~\ref{fusionFramework}, DA strategies for SS can be roughly categorized into three levels: input-level, feature-level, and output-level adaptation.
Input-level adaptation directly reduces distribution gaps in the image space, feature-level adaptation encourages learned representations to be domain-invariant, and output-level adaptation aligns prediction maps or pseudo-labels between the SD and TD.

\noindent\textbf{Input-level adaptation.}
Input-level DA typically relies on style transfer or image translation to make the source and target images more similar.
Many GAN-based approaches have been proposed to transform images from either the SD or TD \cite{benjdira2019unsupervised, tasar2020colormapgan, ji2020generative, cai2022iterdanet, tasar2020standardgan, wittich2021appearance, peng2021full}.
Among them, CycleGAN is one of the most widely used methods. It provides a bidirectional image-to-image translation framework that facilitates the transfer of knowledge from a labeled source domain to an unlabeled target domain \cite{benjdira2019unsupervised, ji2020generative}.
Tasar et al.\ \cite{tasar2020standardgan} further explored realistic scenarios with multiple source domains exhibiting distinct distributions, and subsequently proposed ColorMapGAN, a simplified model that only transfers color information from training to test images to better preserve the geometric structure of RSIs \cite{tasar2020colormapgan}.
In addition to DL-based techniques, classical methods such as the Wallis filter have also been employed for radiometric alignment at the input level \cite{peng2021full}.
However, due to the limited number of training samples and the small spatial extent of image patches, input-level adaptation has received relatively little attention in patchwise RSISS, where learning stable translation models remains challenging.

\noindent\textbf{Feature-level adaptation.}
Feature-level DA aims to learn domain-invariant representations by encouraging the alignment of feature distributions between the SD and TD.
This is typically achieved through metric-based losses combined with backpropagation and can be implemented via adversarial learning, explicit metric alignment, or reconstruction- and disentanglement-based strategies \cite{zhang2021topological, zhang2022graph, zhao2022cross}.
Commonly used metrics include adversarial loss, covariance and parameter losses, maximum mean discrepancy (MMD), contrastive domain discrepancy, Wasserstein distance, and cosine distance \cite{ma2023unsupervised, wang2023fine, lu2021cross, wu2022deep}.

Adversarial-based DA leverages a domain discriminator to encourage the model to extract features that are indistinguishable between the SD and TD, thereby learning domain-invariant and discriminative representations.
Typical adversarial DA networks consist of a feature-extraction subnetwork and a domain-discriminator subnetwork.
For example, the two-branch attention adversarial DA network proposed by Huang et al.\ \cite{huang2022two} employs a dual-branch attention feature extractor to capture spectral–spatial attention features, together with a discriminator containing two classifiers.
Xu et al.\ \cite{xu2023graph} further introduced a graph-guided DA few-shot learning method, which combines graph neural networks and adversarial training after feature extraction to construct a graph-structured domain discriminator.

Metric alignment explicitly constrains the feature distributions of the SD and TD using distance-based objectives.
Liu et al.\ \cite{liu2020class} align the conditional distribution of each class by combining a class-wise domain adversarial neural network with MMD, using pseudo labels of target data during optimization.
PCDM-UDA proposed a multi-view unsupervised DA method that integrates pseudo-class distribution-guided label correction, phase-based domain-invariant features, and trusted prediction to enhance cross-scene HSI classification \cite{gao2025pseudo}.
Optimal transport-based methods learn a transport matrix to map the source distribution to the target distribution and achieve feature alignment by minimizing the transport cost \cite{damodaran2018deepjdot}.
For instance, Zhang et al.\ \cite{zhang2021topological} proposed TsTNet, a topological structure and semantic information transfer network that incorporates topological priors for cross-scene classification.

Reconstruction-based and feature-disentanglement-based DA reduce domain discrepancy by decomposing representations into domain-shared and domain-specific components.
GANs and their variants are widely applied in HSI DA to implicitly enforce reconstruction constraints \cite{he2025hyper}.
Ye et al.\ \cite{ye2024building} form a cross-domain mapping chain by connecting multiple CycleGANs in series; mapping errors are accumulated and backpropagated in each cycle, significantly improving the accuracy of cross-sensor HSI mapping.
Another line of work focuses explicitly on disentangling domain information.
The FDDAN model proposed by Li et al.\ \cite{xin2024feature} extracts domain-shared and domain-specific features through a feature disentanglement network and reconstructs the input to constrain the effectiveness of the separation.
Similarly, the S4DL framework by Chen et al.\ designs a spectral–spatial channel mask to separate domain information, ensuring that alignment between the SD and TD is conducted only on channels that are free from domain bias \cite{feng2024s4dl}.
Beyond HSI, Zhang et al.\ \cite{zhang2020unsupervised} introduced a layer alignment strategy using covariance and parameter losses to mitigate domain shift in high-resolution RSIs.
Wu and Wang \cite{wu2022deep, wang2023fine} proposed a two-stage UDA framework that combines adversarial learning and self-training for optical RSIs.
Lu et al.\ \cite{lu2021cross} presented an end-to-end global–local alignment mechanism with dynamically adjusted adversarial weights, and Ma et al.\ \cite{ma2023unsupervised} incorporated attention mechanisms into GAN-based UDA frameworks to capture long-range dependencies between high-level features from different domains.

\noindent\textbf{Output-level adaptation.}
Output-level DA focuses on aligning prediction maps or pseudo labels to reduce domain shift.
This is typically achieved through adversarial learning \cite{zheng2021entropy, chen2022unsupervised} or self-training \cite{li2022unsupervised, wang2022cross}.
In adversarial learning, a domain discriminator is applied to the prediction space to encourage domain-invariant and confident outputs.
Zheng et al.\ \cite{zheng2021entropy} and Chen et al.\ \cite{chen2022unsupervised} employed entropy-guided adversarial models that emphasize low-confidence regions in the TD during adaptation.
Self-training methods iteratively refine predictions in the TD using a model trained on the SD.
Inspired by DAFormer \cite{hoyer2022daformer}, Li et al.\ \cite{li2022unsupervised} proposed a Transformer-based self-training framework that incorporates gradual class weighting and local dynamic quality estimation to enhance UDA.
In \cite{wang2022cross}, self-training is used to iteratively update pseudo labels and improve cross-domain segmentation performance.
Combining adversarial learning with self-training can further improve DA performance by leveraging high-quality pseudo labels \cite{zhu2023unsupervised, ma2023domain, zhang2021curriculum}.
For example, Ma et al.\ \cite{ma2023domain} introduced a strategy based on local consistency and global diversity metrics to strengthen adaptation in RSIs.
Tong et al.\ \cite{tong2020land} proposed an output-level adaptation method based on pseudo-labelling and a retrieval-based sample selection strategy.

\subsection{Domain generalization (DG)}
Unlike DA, which typically assumes access to labelled SD data and unlabeled TD data, DG tackles the more challenging setting in which models are trained only on one or multiple SD datasets and must generalize to unseen TD data.
DG, as one of the ultimate goals of RS feature extraction, existing methods can be broadly grouped into three categories: (i) data manipulation strategies that generate diverse auxiliary domains or styles \cite{zhang2023single, zhao2023locally, dong2024spectral, cai2025dynamicstyle, iizuka2023frequency, luo2025geosabasa}, (ii) representation learning methods that explicitly enforce domain-invariant feature representations \cite{zhang2023ldgnet, li2025scpnet, qin2024fdgnet, gao2025isdgs, han2024mscdg}, and (iii) general learning strategies that design robust optimization objectives or training pipelines to improve out-of-domain performance \cite{wu2025freemix, wang2024hypersigma, wang2025stylemap, luo2025geosabasa, yaghmour2025sensoragnostic, zhao2025crlmdg, gong2024crossearth}.

\noindent\textbf{Data manipulation.}
In patch-based classification, most existing work focuses on data manipulation. Data manipulation denotes strategies that perturb or extend the SD data distribution to approximate potential domain shifts. 
This can be implemented through various data augmentation techniques, such as texture or style randomization and frequency-based style mixing, or by learnable generators that map SD images to one or more extended domains. 
In contrast to DA, where image translation is guided by real TD data, data manipulation in DG is performed without TD supervision and instead exposes the model to a range of synthetic domain variations to facilitate the learning of domain-invariant representations.
For example, Zhang et al.\ \cite{zhang2023single} designed a morph encoder to learn domain-invariant morphological knowledge during domain expansion, highlighting that DG can be more practical than traditional DA when annotated TD data are unavailable.
Zhao et al.\ \cite{zhao2023locally} proposed a symmetric encoder–decoder to construct extended domains, combined with adversarial contrastive learning to obtain domain-invariant features.
$S^2$ECNet \cite{dong2024spectral} simulates spectral and spatial deviations from the TD via two independent branches, and further introduces a causal alignment module with contrastive learning to extract causal invariant features and mitigate the bias caused by direct feature alignment.
Recent work on dynamic style manipulation further enriches this line by generating structurally constrained auxiliary domains for cross-scene HSI classification \cite{cai2025dynamicstyle}.

Given the persistent shortage of labelled data, data manipulation again emerges as an efficient and straightforward strategy to improve model generalization in tile-based segmentation.
Specifically, Iizuka et al.\ \cite{iizuka2023frequency} proposed FOSMix, a frequency-based augmentation technique that blends image styles in the frequency domain while preserving semantic content, thereby enhancing domain diversity.
Liang et al.\ \cite{liang2024ccdr} introduced CCDR, an SDG method that employs randomized texture and style transformations together with a class-aware consistency constraint, achieving strong generalization with a simple training pipeline.

\noindent\textbf{Domain-invariant feature representation.}
A second line of work targets DG by explicitly learning domain-invariant feature representations.
In HSI classification, language-aware DG networks such as LDGnet \cite{zhang2023ldgnet} leverage large-scale image-text models with dual-stream encoders, aligning visual and linguistic features in a shared semantic space via supervised contrastive learning to obtain cross-domain-invariant representations.
Semantic-aware co-parallel architectures such as SCPNet \cite{li2025scpnet} employ dual-branch spatial-spectral and multiscale modules, along with an optimized semantic space, to collaboratively extract domain-invariant features.
Frequency- and geometry-based models such as FDGNet \cite{qin2024fdgnet} further disentangle frequency components and exploit data geometry on Riemannian manifolds to suppress domain-specific artefacts while preserving discriminative structure, whereas invariant semantic DG shuffle networks refine the semantic space by shuffling and aligning semantic components across domains \cite{gao2025isdgs}.
At the multi-source level, MS-CDG \cite{han2024mscdg} combines data-aware adversarial augmentation with model-aware multi-level diversification and distribution-consistency alignment to jointly encourage domain-invariant representation learning across heterogeneous sources.
Apart from HSI classification, similar principles of learning style-robust and semantically consistent representations are also adopted implicitly in RGB and multispectral images, where they are usually integrated into the backbone design or loss formulation and are therefore discussed together with general DG training strategies below rather than as a separate category.

\noindent\textbf{General learning strategy.}
Several works focus on general learning strategies that couple data- and feature-level DG in a unified training pipeline \cite{wang2024hypersigma, zhang2025spectralx, wu2025freemix, wang2025stylemap, luo2025geosabasa, yaghmour2025sensoragnostic, gong2024crossearth, zhao2025crlmdg}.
StyleMap \cite{wang2025stylemap} regularizes both feature-space geometry and style diversity by combining style mapping with a prototypical contrastive loss, providing a theory-driven DG framework.
A second group of works builds on vision or geospatial VFMs: GeoSA+BaSA \cite{luo2025geosabasa} and sensor-agnostic DG \cite{yaghmour2025sensoragnostic} couple PEFT with batch-wise style augmentation or generative pre-training to obtain sensor- and scene-agnostic representations, 
while CrossEarth \cite{gong2024crossearth} targets remote-sensing DG by combining an Earth-Style Injection augmentation pipeline with multi-task training that jointly learns semantic segmentation and masked image modeling on a shared DINOv2 backbone, and it validates this design on an RSDG benchmark covering 28 cross-domain settings, showing strong cross-scene generalization.
Beyond generic land-cover mapping, CRLMDG \cite{zhao2025crlmdg} introduces causal representation learning and multi-target consistency for fine-grained landslide mapping under complex domain shifts.
HyperSIGMA \cite{wang2024hypersigma} and SpectralX \cite{zhang2025spectralx} share a Transformer core and self-supervised masked reconstruction pretraining on large spectral corpora to learn transferable spatial–spectral representations. 
They explicitly address HSI redundancy via specialized attention/tokenisation and add lightweight adaptation components to support downstream classification and generalize to unseen regions, seasons, and tasks.
FreeMix \cite{wu2025freemix} extends DG to an open-vocabulary setting by combining entity-mask mixing with RS-specific self-supervised initialization to jointly handle unseen domains and unseen classes.

\subsection{Self-supervised learning (SSL)}
SSL, a powerful model pre-training method, enables models to extract useful knowledge from unlabeled data by designing pretext tasks whose supervision signals are automatically derived from the data itself \cite{wang2025non}.
After training on large-scale unlabeled RSIs, the learned representations can be adapted to downstream classification or segmentation tasks using only a small number of labelled samples.
In RS, SSL methods for SS can be broadly grouped into two primary sub-directions: (i) self-predictive generative learning methods, which reconstruct or predict masked/perturbed content \cite{wang2024hypersigma, mou2017unsupervised, mei2019unsupervised, ibanez2022masked, scheibenreif2023masked, DBLP00953, li2021semantic, sun2022ringmo, cong2022satmae, cai2024consistency, liu2023rethinking, cao2023transformer}, and (ii) discriminative contrastive learning methods, which learn to distinguish similar from dissimilar representations \cite{yu2023hyperspectral, liu2020deep, chang2024unsupervised, DBLP00953, app11188670, hou2021hyperspectral, guan2022cross, li2022global, muhtar2022index, li2021semantic, cao2023transformer, rs13163275, dong2023spatial}.

\noindent\textbf{Self-predictive generative learning.}
Early SSL for patchwise SS mainly relied on AE frameworks.
AE models were extended to CNN architectures, where Conv–Deconv networks learn spectral–spatial features in an unsupervised manner and then require only a small number of labels to achieve strong supervised classification performance \cite{mou2017unsupervised, mei2019unsupervised}.
However, the compressed features obtained by AEs are not always optimal for downstream tasks, motivating richer self-predictive objectives.
Subsequent works designed pretext tasks that explicitly exploit the inherent relationships within RSIs, such as image inpainting, transformation prediction, and solving jigsaw puzzles \cite{li2021semantic}.
These innate relationship prediction tasks encourage the model to understand local structural coherence, but often focus on narrow and specific cues, limiting their generality.

Masking-based learning has therefore gained increasing popularity \cite{he2022masked}.
In this paradigm, parts of the input are intentionally hidden, and the model is trained to reconstruct the missing content, which helps capture both local and global dependencies and uncover contextual and structural relationships in the data.
Originally proposed in denoising AEs \cite{vincent2008extracting}, masking-based SSL evolved through BEiT \cite{chen2020generative} and matured in masked autoencoders (MAE) \cite{he2022masked}.
In patchwise classification, MAEST \cite{ibanez2022masked} was the first to integrate the masking–reconstruction strategy of MAE with spectral–spatial feature extraction.
HyperSIGMA \cite{wang2024hypersigma} pre-trained with MAE on the large HyperGlobal-450K dataset. 
It separately pre-trains spatial and spectral subnetworks, enabling strong cross-scene transfer across diverse high- and low-level HSI tasks.
To further explore the potential of SSL for HSI analysis, MSST \cite{scheibenreif2023masked} constructed a large EnMAP-based dataset and used masked image reconstruction to pretrain Transformer models, yielding robust spectral–spatial representations.
For tilewise SS, self-predictive SSL is also widely used.
Methods such as RingMo and SatMAE design RS-specific masking patterns and reconstruction targets, enabling ViT-like backbones to learn generalizable features from large-scale multi-sensor RSIs \cite{sun2022ringmo, cong2022satmae}.
Recent consistency-regularized masking strategies further improve robustness by enforcing prediction consistency under different masking configurations \cite{cai2024consistency, liu2023rethinking}.

\noindent\textbf{Discriminative contrastive learning.}
Discriminative SSL aims to learn the similarities and differences between data samples, typically by constructing positive and negative pairs and optimizing a contrastive objective.
In patchwise classification, contrastive learning has become a mainstream approach for unsupervised feature extraction.
Data augmentation is the most common means for creating positive/negative pairs \cite{app11188670, hou2021hyperspectral}.
Liu et al.\ \cite{liu2020deep} employed a multi-view approach by grouping HSI bands to form sample pairs.
Cao et al.\ \cite{DBLP00953} used two separate AEs to generate sample pairs and built a ProtoNCE-based contrastive learning model.
XDCL \cite{guan2022cross} introduced a cross-domain discrimination task that leverages both spatial and spectral information to extract shared representations.
Graph-based methods further exploit the relational structure of HSI data: graph augmentation modules generate paired graphs for contrastive learning in GCNs, leading to domain-invariant and topology-aware features \cite{yu2023hyperspectral, chang2024unsupervised}.

In tilewise SS, contrastive SSL has also been widely adopted.
Earlier works \cite{li2021semantic, rs13163275} trained encoders using classical instance-discrimination frameworks, whereas later methods proposed multi-granularity contrastive objectives to capture information at different levels.
Li et al.\ \cite{li2022global} jointly extracted local and global representations, Muhtar et al.\ \cite{muhtar2022index} focused on both pixel-level and image-level representations, and Dong et al.\ \cite{dong2023spatial} combined instance-level and semantic-level contrast.
Despite their effectiveness, contrastive methods generally rely on complex data augmentation pipelines and meticulous design of positive and negative pairs, which can make them less flexible and harder to tune than purely masking-based approaches.

To exploit the complementary strengths of generative and discriminative SSL, hybrid strategies have been proposed.
Cao et al.\ \cite{cao2023transformer} introduced a Transformer-based framework that integrates contrastive and masked learning, enabling both pixel-level feature learning and global spectral–spatial representations.
This hybrid design consistently outperforms models that rely on a single learning paradigm.

\subsection{Semi-supervised learning (SeL)}
SeL provides a balanced and effective alternative to purely supervised or purely unsupervised schemes by integrating task-specific knowledge from supervised learning and task-agnostic knowledge from unsupervised learning.
SeL is particularly attractive in RS because labeled RSIs are scarce and expensive to acquire, whereas unlabeled data are abundant \cite{chen2020big}.
Existing SeL methods for SS can be broadly organized into three strategy families: (i) self-training and graph-based pseudo-label propagation \cite{dong2022weighted, liu2020cnn, wu2017semi, seydgar2022semisupervised, 10329942, wang2022semi, lu2022simple}, (ii) joint optimization of supervised and unsupervised losses \cite{liu2024universal, liu2017semi, hang2020adversarial, huang2023spectral}, and (iii) consistency regularization  \cite{huang2024decouple, 10309935, lu2022simple, li2021semisupervised, jin2024dynamic, wang2020semi, wang2021ranpaste, lv2024advancing, chen2024category, chen2022semi, guo2023semisupervised, kang2021picoco, rs14040879, chen2023semi}.

\noindent\textbf{Self-training and graph-based pseudo-label propagation.}
Self-training, also referred to as pseudo-labelling, is one of the most widely used SeL strategies.
An initial classifier predicts labels for unlabeled samples, and these predictions are iteratively refined and reused as pseudo labels for further training \cite{wu2017semi, seydgar2022semisupervised, 10329942, wang2022semi, lu2022simple}.
Self-training is often coupled with adaptive thresholding to select high-confidence pseudo labels
For example, ICNet \cite{wang2022semi} introduces an iterative contrastive network that alternates updates between paired networks to enhance pseudo-label quality and segmentation performance.
In graph-based patchwise classification, works \cite{dong2022weighted, liu2020cnn} extend this idea by using superpixels or graph nodes to propagate labels, which can also be treated as pseudo-labels.
Combining pseudo-labels with real labels helps mitigate the problem of limited annotated samples, but overall performance is highly sensitive to the quality of pseudo-labels, making their refinement an ongoing challenge.

\begin{figure}[t]
\centering
\includegraphics[width=3.3 in]{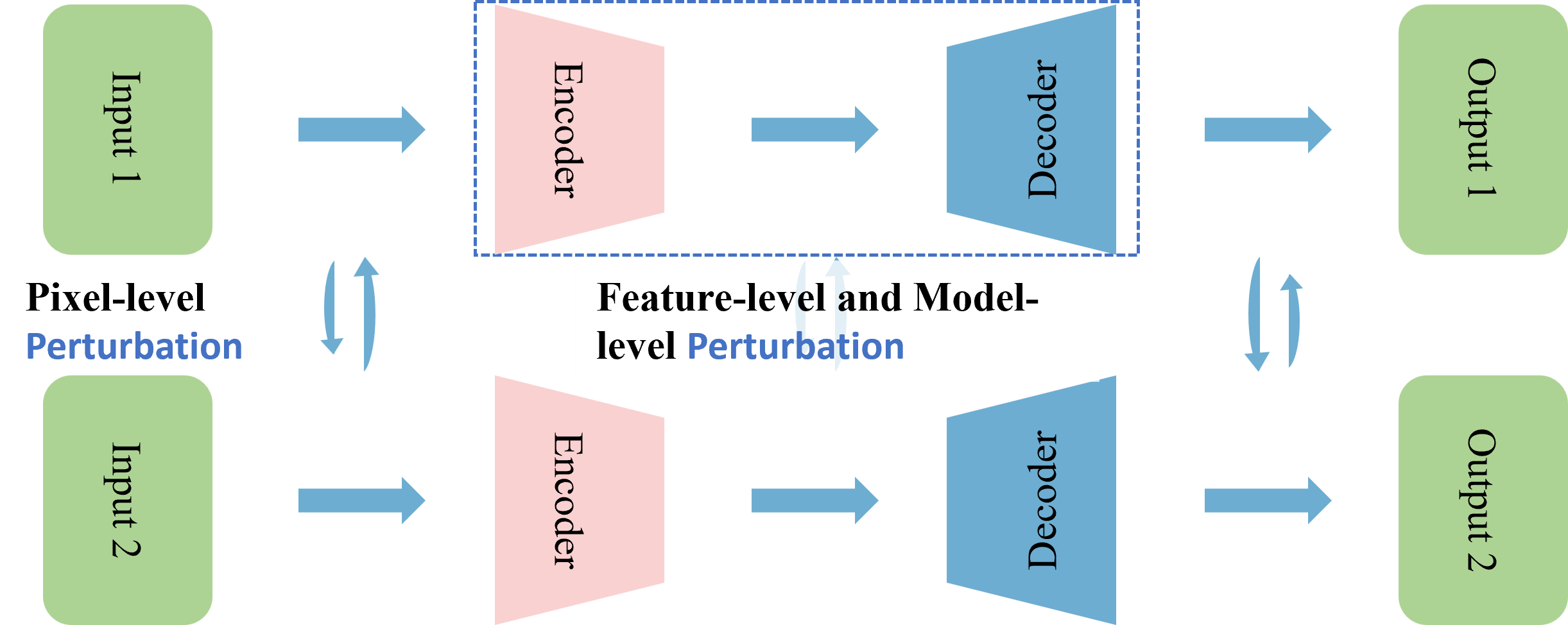}
\caption{\textbf{Unified remote sensing image semantic segmentation
framework incorporating perturbation at multiple levels.}}
\label{perturbationFramework}
\end{figure}

\noindent\textbf{Joint optimization of supervised and unsupervised losses.}
A flexible framework for constructing SeL models is to combine supervised and unsupervised loss functions in a unified objective.
Such frameworks allow models to learn both task-specific and task-agnostic knowledge, effectively leveraging unlabeled data to improve decision-making \cite{liu2024universal, liu2017semi, hang2020adversarial, huang2023spectral}.
For HSI classification, Liu et al.\ \cite{hang2020adversarial} proposed a dual-task model in which a generator and discriminator engage in unsupervised adversarial training across modalities, while the encoder of the generator, equipped with a classification head, is trained in a supervised manner.
Huang et al.\ \cite{huang2023spectral} further advanced this idea by combining contrastive and supervised losses into a reconstruction loss via additional branches, encouraging the network to learn representations that are simultaneously reconstructive, discriminative, and robust.
Related universal semi-supervised strategies couple supervised objectives with unsupervised contrastive losses, striking a favourable balance between training speed and effectiveness across multiple HSI datasets.

\noindent\textbf{Consistency regularization.}
Consistency regularization constitutes a third line of SeL research and is particularly prevalent in tilewise SS.
The central idea is to enforce prediction consistency under multiple perturbations of the same input, thereby improving generalization and reducing reliance on large labelled datasets \cite{li2021semisupervised}.
Perturbations can be applied at the input, feature, and model levels, as illustrated in Figure~\ref{perturbationFramework}.
Most existing methods focus on pixel-level perturbations and employ random image augmentations such as color jittering \cite{wang2020semi}, random paste \cite{wang2021ranpaste}, cutout, edge enhancement, grayscale conversion, and blurring \cite{lu2022simple, jin2024dynamic}.
More advanced strategies further incorporate pseudo-label selection \cite{qi2023pics}, CutMix-based mixing \cite{lv2024advancing}, or combinations of multiple transformations to promote the learning of domain-invariant features \cite{chen2024category}.
In contrast, relatively few studies consider feature- or model-level perturbations in isolation.
Chen et al.\ \cite{chen2022semi} introduced random drop and noise in the feature-map domain, while Li et al.\ \cite{li2021semisupervised} proposed several types of random feature perturbations within a GAN framework to optimize a consistency loss.
Model-level perturbation techniques, such as mean teacher and cross pseudo supervision, introduce variations in model weights or structure to improve robustness.
In practice, applying perturbations at multiple levels simultaneously has proved more effective for learning invariant representations \cite{huang2024decouple, 10309935, guo2023semisupervised}.
For instance, PiCoCo \cite{kang2021picoco} and ClassHyPer \cite{rs14040879} incorporate both feature- and sample-level consistency regularization, while Semi-FCMNet \cite{chen2023semi} combines data augmentation with MT across input, model, and feature levels to maintain consistency and amplify salient features.

Overall, patchwise classification has primarily used self-training and joint optimization, whereas tilewise segmentation has driven the development of richly regularized frameworks, offering promising directions for future patchwise SeL designs.

\subsection{Weakly supervised learning (WSL)}

Coarse annotations, such as image-level labels, bounding boxes, point annotations, and scribble annotations, are significantly easier to obtain than pixel-level labels.
Weakly supervised learning semantic segmentation (WSLSS) methods offer a practical solution for performing SS under weak supervision, mitigating the challenges posed by the lack of dense pixel annotations.
The general framework of WSLSS involves two steps: generating pseudo-masks from coarse annotations and then training the model using standard SS techniques \cite{zhou2018brief}.

Among all WSL types, image-level WSL is the most widely studied and challenging.
The foundational idea originates from the work of Zhou et al. \cite{zhou2016learning}, who observed that CNNs exhibit strong localization capabilities even when trained solely on image-level labels.
Their class activation mapping method became a standard technique for generating pseudo-label seeds in WSSS \cite{fu2018wsf}.
Subsequent studies addressed issues related to the noise and incompleteness of initial pseudo labels and spatial context \cite{yang2023iter, CAO2022157, fang2022improved, li2021effectiveness, chen2020spmf, iqbal2020weakly}.
For instance, Yang et al. \cite{yang2023iter} proposed ITER to bridge image-level tags and pixelwise HSI prediction via a two-stage pipeline. CAM-based pseudo-labels are progressively refined using spectral/spatial activation and spectral–spatial alignment losses, after which a highly enhanced Transformer with high-frequency-aware self-attention learns boundary-detailed dense maps.
Javed et al. \cite{iqbal2020weakly} proposed a weakly supervised DA approach for built-up region segmentation, incorporating a detection network that leverages image-level labels to support DA.

Other forms of weak supervision have also been applied to various RSISS tasks.
NFANet \cite{lu2022nfanet} proposed a neighbour sampler to utilize point-level labels for water body extraction.
Wei et al. \cite{9372390} introduced a scribble-based weak-supervision method for road surface extraction that combines road label propagation with holistically nested edge detection to generate training masks.
Li et al. \cite{LI2022244} leveraged low-resolution land-cover products as weak supervision signals and proposed a low-to-high framework for large-scale, high-resolution land-cover mapping.

In summary, while WSSS methods show promise in reducing annotation burdens, further developments are needed to adapt them to the unique characteristics of RSIs.
Future work should prioritize improving boundary detection, addressing scale and class imbalance, and integrating modern techniques such as multimodal fusion and SSL to enhance the generalizability and accuracy of WSL in RSISS.

\begin{figure}[t]
\centering
\subfloat[\textbf{Independent and cross-gated mechanism.}]{\includegraphics[width=1.5 in]{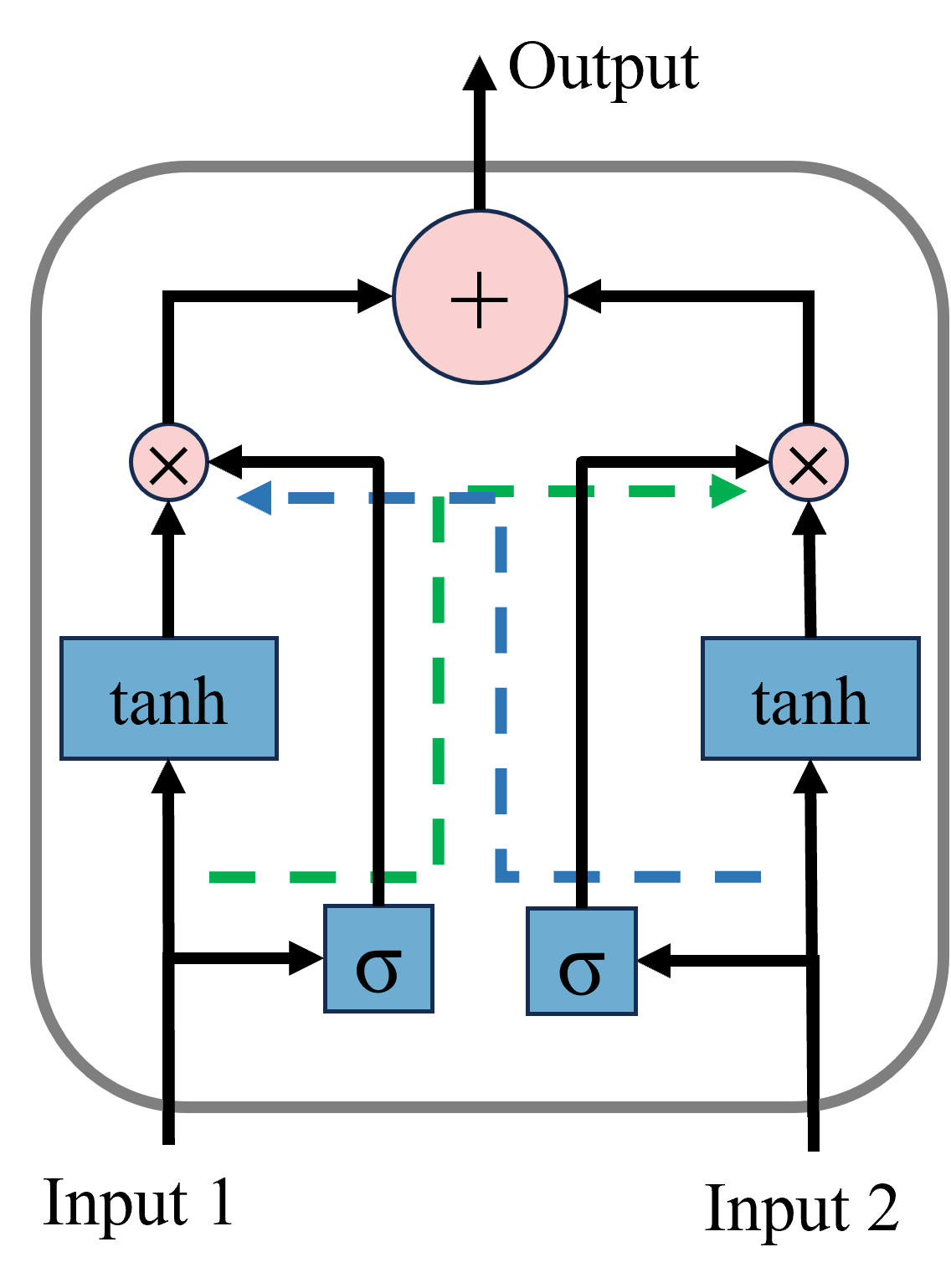}}
\hspace{6mm}
\subfloat[\textbf{Complementary gated mechanism.}]{\includegraphics[width=1.5 in]{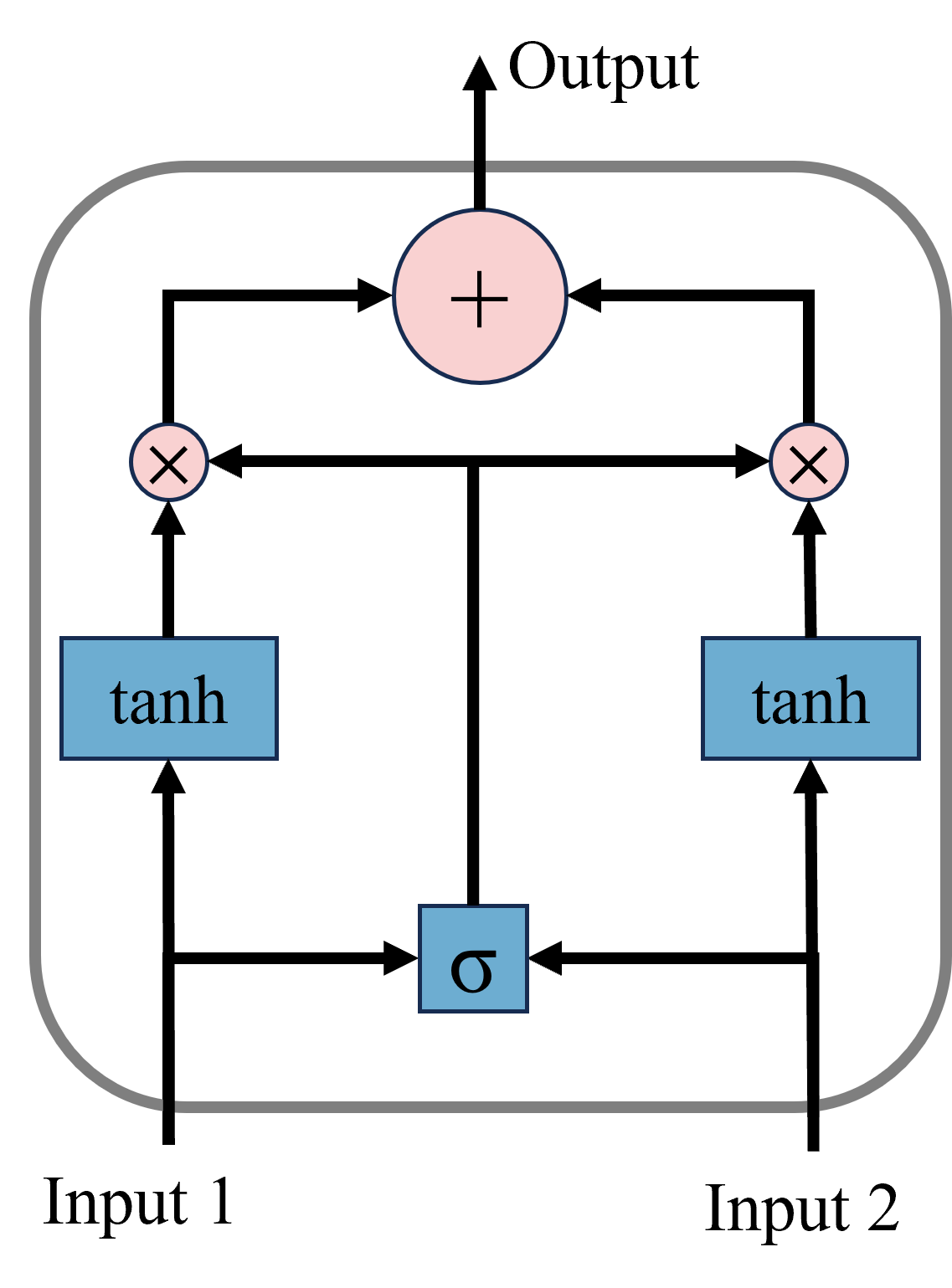}}
\caption{\textbf{Comparison of gated fusion mechanisms for feature modulation. (a) Independent and cross-gated mechanism, where each input generates gating weights via sigmoid activation to modulate either its own features or those of the other branch. (b) Complementary gated mechanism, where a shared gating signal regulates both inputs to encourage complementary feature integration. The \textit{tanh} blocks denote nonlinear feature transformation, and $\sigma$ denotes sigmoid-based gating.}}
\label{GatedUnit}
\end{figure}

\section{Information Fusion Techniques}
\label{ImageFusion}

\begin{table*}[t]
\caption{\textbf{An overview of information fusion strategies for remote sensing image semantic segmentation. Based on the surveyed literature, existing feature fusion techniques are further classified into several subclasses.}}
\label{allMultiReference}
\centering
\small
\renewcommand{\arraystretch}{1.25}
\setlength{\tabcolsep}{4pt}

\begin{tabular}{p{3.6cm}p{6.5cm}p{6.5cm}}
\toprule
\textbf{Feature fusion techniques} &
\textbf{Patch-based feature fusion reference} &
\textbf{Tile-based feature fusion reference} \\
\midrule

\multicolumn{3}{c}{\textbf{Linear operation fusion}} \\
\midrule

Single-point fusion &
Single-branch structure: \cite{paisitkriangkrai2015effective}, \cite{du2021multisource}, \cite{yang2022single}, \cite{yu2022capvit}, \cite{dong2024fusion}, \cite{zhao2020joint} \par
Symmetric structure:  \cite{feng2019multisource}, \cite{chen2017deep}, \cite{li2018hyperspectral}, \cite{li2020robust}, \cite{xue2022self}, \cite{CAI2023119655}, \cite{li2020multimodal}, \cite{zhao2022fractional}, \cite{liu2025SHNet} \par
Asymmetric structure: \cite{guo2022dual}, \cite{xu2017multisource}, \cite{roy2022hyperspectral} &
Single-branch structure: \cite{kampffmeyer2016semantic}, \cite{yue2019treeunet}, \cite{diakogiannis2020resunet}, \cite{xie2022glaciernet2}, \cite{liu2017dense} \par
Symmetric structure:  \cite{audebert2016semantic}, \cite{liu2023joint} \par
Asymmetric structure: \cite{sherrah2016fully} \\
\midrule

Multi-point fusion &
Symmetric structure: \cite{hang2020classification}, \cite{song2022discriminative}, \cite{zhang2023cross}, \cite{zhao2021fractional}, \cite{Zhang2026Multimodal},  \cite{li2024hypermlp} \par
Asymmetric structure: \cite{gao2023amsse}, \cite{zhang2022multimodal}, \cite{du2023hyperspectral}, \cite{ni2024mhst} &
Symmetric structure:  \cite{audebert2018beyond}, \cite{peng2019densely},  \cite{wang2024multisenseseg}, \cite{he2023sfaf} \\
\midrule

\multicolumn{3}{c}{\textbf{Nonlinear interaction fusion}} \\
\midrule

Attention mechanism fusion &
Single cross-attention: \cite{mohla2020fusatnet}, \cite{dong2022multibranch}, \cite{xiu2022mage}, \cite{wang2022am3net}, \cite{roy2023multimodal}, \cite{yao2023extended}, \cite{zhao2022joint}, \cite{zhang2023hyperspectral}, \cite{wang2023mutually}, \cite{roy2024cross} \par
Mutual cross-attention:\cite{xue2022deep}, \cite{wang2024ms2canet}, \cite{li2022triplet}, \cite{li2021emfnet}, \cite{fang2021s2enet}, \cite{zhang2021mutual}, \cite{wang2022multi}, \cite{feng2023dshfnet}, \cite{zhang2022local},  \cite{li2023mixing}, \cite{zhang2024cross}, \cite{gao2025msfmamba} &
Single cross-attention: \cite{ren2022dual} \par
Mutual cross-attention: \cite{zhang2020hybrid}, \cite{yang2021attention}, \cite{ferrari2021integrating}, \cite{ma2022crossmodal}, \cite{sun2021deep}, \cite{li2022mcanet}, \cite{ma2024multilevel} \\
\midrule

Gated mechanism fusion &
Independent gated mechanism: \cite{li2023mixing} \par
Complementary gated mechanism:  \cite{li2020collaborative}, \cite{zhang2023sar}, \cite{li2024automatic} &
Independent gated mechanism: \cite{huang2019automatic}, \cite{yue2024bclnet} \par
Complementary gated mechanism:  \cite{zhou2021cegfnet}, \cite{hosseinpour2022cmgfnet}, \cite{geng2023multisource}, \cite{wei2024mgfnet} \par
Cross-gated mechanism: \cite{kang2022cfnet} \\
\midrule

Reconstruction mechanism fusion &
Self-reconstruction: \cite{ding2022global}, \cite{hong2020deep}, \cite{zhang2021information}, \cite{lu2023coupled}, \cite{zhang2024multimodal} \par
Cross-reconstruction: \cite{wu2021convolutional}, \cite{zhang2018feature}, \cite{pande2023self} &
Self-reconstruction: \cite{hong2020multimodal}, \cite{chen2023incomplete}\\
\midrule

Alignment mechanism fusion &
\cite{hong2020more}, \cite{Hang2022CrossModality} &
\cite{hong2020multimodal}, \cite{li2023progressive}, \cite{li2023aligning}, \cite{gupta2025mosaic}, \cite{chen2025mdca}, \cite{li2025semantic} \\
\bottomrule
\end{tabular}
\end{table*}

With the increasing availability of multimodal data, attention has been shifting toward multi-source RSI analysis.
Beyond the feature extraction strategies discussed in the previous section, the fusion module is a critical component for capturing complementary features from multimodal images.
Accordingly, a DL-based fusion taxonomy is introduced that categorizes fusion techniques into simple linear-operation fusion and complex nonlinear-interaction fusion.
Linear operation fusion integrates multimodal information through concatenation or linear weighting, whereas nonlinear interaction fusion achieves specific fusion objectives via more complex operations.
This taxonomy provides a more nuanced study direction for future data fusion research.
It is worth noting that these strategies are not only essential for multimodal data but also widely applicable to multi-branch fusion settings in unimodal SS. Meanwhile, these strategies cover a spectrum ranging from pixel-level classification to holistic image-level segmentation.
Table \ref{allMultiReference} summarises the main patch-based and tile-based feature fusion for semantic segmentation
models.

\subsection{Linear operation fusion}

Linear operation fusion is a simple yet effective method widely used in multimodal fusion approaches, which consists of elementary arithmetic and concatenation.
As illustrated in Figure~\ref{fusionFramework}, the fusion operation can be implemented at different stages of the model pipeline, namely: pixel-level fusion \cite{kampffmeyer2016semantic, yang2022single}, feature-level fusion \cite{feng2019multisource, audebert2016semantic}, and decision-level fusion \cite{zhao2020joint, liu2017dense}. Moreover, the fusion operation can be performed once or multiple times, namely: single-point fusion and multi-point fusion. 
As the name suggests, single-point fusion only performs one fusion operation, and multi-point fusion conducts multiple fusions during forward propagation.

In the early stages of multimodal fusion, raw data is typically channel-wise concatenated directly to maintain compatibility with traditional unimodal models \cite{kampffmeyer2016semantic, paisitkriangkrai2015effective, du2021multisource, yang2022single, yu2022capvit, dong2024fusion, yue2019treeunet, diakogiannis2020resunet, xie2022glaciernet2}. 
For instance, Yang et al. \cite{yang2022single} stack HSI and LiDAR channels at the input and rely on a single-stream backbone to capture cross-modal interactions through standard convolutional feature extraction realized with group convolutions. 
Dong et al. \cite{dong2024fusion} produce a single fused representation from GF-5 and Sentinel-2B at the pixel level and then perform downstream lithology classification using a ViT encoder followed by graph-based modeling in ViT-DGCN.
Kampffmeyer et al. \cite{kampffmeyer2016semantic} first demonstrated the feasibility of tilewise RSISS using a vanilla FCN architecture, in which pixel-level concatenation of the true orthophoto, digital surface model (DSM), and normalized DSM was performed.
Yue et al. \cite{yue2019treeunet} append DSM-derived depth cues to the optical channels and feed the resulting stacked tensor into a U-Net-style encoder–decoder composed of conventional convolutional blocks for SS. 
Xie et al. \cite{xie2022glaciernet2} stacking heterogeneous geospatial sources as the network input, and they learn fused representations using classical CNN segmentation backbones such as SegNet with VGG16-style feature extraction, complemented by DeepLab-inspired components in their hybrid architecture. 

In a single-branch formulation, the model keeps one primary predictor, while the auxiliary modality is introduced only as an external constraint or prior during post-processing, rather than through an explicit parallel feature-extraction branch. In this spirit, Liu et al.  \cite{liu2017dense} use an FCN as the sole dense predictor on optical imagery and inject LiDAR-derived cues as additional unary information within a higher-order CRF to refine the final labelling. 
Likewise, Zhao et al. \cite{zhao2020joint} treat the CNN output as the main decision signal and incorporate DSM-driven pixel affinities in a hierarchical random-walk layer to regularize predictions and improve spatial consistency. 
This design is attractive because it preserves a compact, unimodal-compatible backbone and typically reduces parameter growth by avoiding a second deep branch, while still leveraging complementary modalities through plug-in structured inference, which can enhance boundary coherence and suppress noise without redesigning the main network.

%
However, the network must learn cross-modal interactions from raw stacked channels, making it more sensitive to dynamic-range mismatch, modality-specific noise, and residual misregistration, and it does not explicitly model nonlinear inter-modal relationships.
Feature-level fusion integrates extracted features from different modalities to form a more comprehensive representation, enabling deeper fusion. 
The branches used to extract multimodal features can either share the same structure or be customized to the data types, i.e., symmetric or asymmetric.
Single-point symmetric fusion is the most common form of patchwise multimodal fusion \cite{chen2017deep, li2018hyperspectral, li2020robust, xue2022self, CAI2023119655, li2020multimodal, zhao2022fractional}.
In \cite{xue2022self} and \cite{CAI2023119655}, a shared encoder was employed during training, where contrastive and clustering losses from each branch were aggregated.
Feature interactions between branches were learned using backpropagation.
Du et al. \cite{du2021multisource} trained a multimodal graph and CNN framework using two unsupervised loss functions and applied an SVM classifier to the unified fused feature representation.
In addition, CapViT \cite{yu2022capvit} integrated CapsNet with a Transformer encoder to enable long-range global feature fusion from multi-scale patches.
FrIT \cite{zhao2022fractional} employed a fractional Fourier image transformer to capture global contextual and sequential information through a linear fusion strategy.

Due to increased model complexity, only a few works focus on leveraging single-point fusion in tile-based segmentation \cite{audebert2016semantic, liu2023joint}.
Multimodal information fusion at a single location often requires a large amount of training data to achieve optimal performance.
To address this limitation, many studies have adopted multi-point linear fusion strategies to enhance modality interaction and improve the efficiency of complementary information extraction \cite{hang2020classification, song2022discriminative, zhang2023cross, zhao2021fractional, Zhang2026Multimodal, li2024hypermlp, audebert2018beyond, peng2019densely, wang2024multisenseseg, he2023sfaf}.
In patchwise classification, Zhang et al. \cite{zhang2023cross} aligned the source and target domains before applying a classification network in combination with multi-point linear fusion, achieving strong cross-scene performance.
Zhao et al. \cite{zhao2021fractional} introduced Octave convolution and fractional Gabor convolution to preserve multisource, multiscale, and multidirectional features during feature extraction.
Zhang et al. \cite{Zhang2026Multimodal} performed concatenation and addition operations on features derived from a Transformer encoder to achieve effective fusion.
In tilewise SS, FuseNet \cite{audebert2018beyond} demonstrated that early summation fusion improves classification accuracy by leveraging the complementary nature of multimodal data through joint feature learning.
Other works followed this early feature-fusion strategy, incorporating additional enhancements.
Peng et al. \cite{peng2019densely} and He et al. \cite{he2023sfaf} combined early feature fusion with dense connections, atrous convolution, and attention mechanisms.
More recently, MultiSenseSeg \cite{wang2024multisenseseg} introduced modality-specific experts to extract consistent features and reduce inter-modal differences before concatenation for downstream encoding and decoding.

The above approaches employ symmetric network architectures for all modalities.
Designing non-symmetric networks tailored to different modalities is considered more effective for handling heterogeneous multimodal data \cite{guo2022dual, du2023hyperspectral}.
Guo et al. \cite{guo2022dual} emphasized the unique characteristics of spectral and spatial modalities, advocating for distinct feature extraction strategies.
Du et al. \cite{du2023hyperspectral} concatenated pixel-level and feature-level representations to extract both local and global features from HSI and LiDAR data.
Several works \cite{sherrah2016fully, gao2023amsse, zhang2022multimodal, ni2024mhst, xu2017multisource} employed two different network structures to separately extract spectral features from HSIs and spatial features from LiDAR data under both single-point and multi-point linear fusion frameworks.
Sherrah et al. \cite{sherrah2016fully} fused a pretrained CNN and FCN within the encoder to generate full-resolution labelling without downsampling.
AMSSE-Net \cite{gao2023amsse} implemented an involution operation for spectral feature characterization and applied five linear operation combinations to fuse multimodal features.
MHST \cite{ni2024mhst} processed fused multimodal features through CNNs and Transformer blocks to extract global spectral and local spatial information, respectively.

\subsection{Nonlinear interaction fusion}

Multimodal feature fusion provides a more comprehensive feature representation.
However, simple linear fusion of individually extracted features can lead to redundancy and increase the risk of overfitting \cite{mohla2020fusatnet}.
To address redundant information generated by modality imbalance and rigid feature stacking, various nonlinear interaction fusion methods have been proposed.
These methods incorporate attention, gating, reconstruction, and alignment mechanisms to facilitate deeper feature interactions and enhance SS performance.

\subsubsection{Attention mechanism fusion}

Multimodal SS approaches frequently use attention mechanisms to assign weights to different spatial or spectral regions, enabling the model to focus on informative components across modalities.
As shown in Figures~\ref{implicitAttention} and~\ref{explicitAttention}, when attention weights are derived from other modalities, a cross-attention module is constructed.
Based on the structure of fusion models, attention mechanism fusion can be categorised into single cross-attention and mutual cross-attention approaches.
This design reflects the varying importance and contribution of each modality to the overall task.

For the single cross-attention model, weights are generated from a single modality and applied to guide feature extraction in another modality.
This approach is based on the understanding that different modalities carry varying levels of information. Exploiting privileged information from the dominant modality can improve model performance, whereas the other modality may introduce noise or redundancy. \cite{mohla2020fusatnet, dong2022multibranch, xiu2022mage, wang2022am3net, roy2023multimodal, yao2023extended, zhao2022joint, zhang2023hyperspectral, wang2023mutually, roy2024cross, ren2022dual}.
CNN remains a widely used backbone in cross-attention modules due to its efficiency in capturing local spatial structures and contextual information.
FusAtNet \cite{mohla2020fusatnet} was the first to use a cross-attention mechanism in HSI classification where the attention map derived from LiDAR data was used to emphasise the spatial features of HSI.
Dong et al. \cite{dong2022multibranch} also generate an attention mask from the LiDAR features and then reuse this mask to modulate both streams, so that elevation cues are injected while spatial structures in HSI are explicitly enhanced.
Ren et al. \cite{ren2022dual} propose a dual-stream HRNet where SAR and optical images are encoded separately, and SAR features are injected into the optical stream at the end of each HRNet stage via cross-modal fusion, providing complementary, all-weather cues to improve dense land-cover labelling.
Subsequent works \cite{xiu2022mage, wang2022am3net, zhang2023hyperspectral} adopted similar strategies to reinforce HSI spatial representations.

Recent studies employed pure Transformer encoders \cite{roy2023multimodal, yao2023extended, wang2023mutually, hu2022hyperspectral} or CNN-Transformer hybrid architecture \cite{zhao2022joint} for multimodal feature extraction.
The key distinction among these implementations lies in how the query, key, and value components are drawn from different modalities to establish cross-attention.
For example, ExViT \cite{yao2023extended} uses cross-modality attention and all-token fusion, enabling information to flow across modalities at multiple levels and yielding more discriminative, synergistic multimodal representations than single-modality baselines.
Zhao et al. \cite{zhao2022joint} propose a hierarchical CNN–Transformer fusion framework for HSI–LiDAR joint classification. CNNs extract spatial–spectral and elevation features, transformers model long-range dependencies via tokenization, and a cross-token attention fusion encoder integrates both modalities to boost land-cover recognition.

In mutual cross-attention methods, attention weights are derived from all participating modalities.
This approach assumes that each modality carries important discriminative information that collaboratively enhances the overall model performance.
The discriminative information is regard as prior in each modality to assist feature extraction in the other modality.
Such cross-attention methods are widely adopted in both patchwise classification \cite{xue2022deep, wang2024ms2canet, li2022triplet, li2021emfnet, fang2021s2enet, zhang2021mutual, wang2022multi, feng2023dshfnet, zhang2022local, li2023mixing, zhang2024cross, gao2025msfmamba} and tilewise segmentation \cite{zhang2020hybrid, yang2021attention, ferrari2021integrating, ma2022crossmodal, sun2021deep, li2022mcanet, ma2024multilevel}.
For example, in patchwise classification, a spectral-spatial enhancement module can simultaneously strengthen spatial features of HSI using LiDAR data and reinforce LiDAR features using spectral information of HSI \cite{li2021emfnet, wang2024ms2canet}.
S$^2$ENet \cite{fang2021s2enet} introduces S$^2$EM with two complementary cross-modal modules: SAEM computes global spatial affinities from LiDAR to reweight HSI features, while SEEM computes channel-wise relations from HSI to recalibrate LiDAR channels, enabling separate extraction of spatial and channel information before fusion.

Another approach involves extracting features from each modality separately to enhance the fused modality or using the fused feature to improve the individual modalities. \cite{zhang2022local, sun2021deep}. 
For instance, TSDN \cite{li2022triplet} concatenates HSI spatial cues and LiDAR elevation and applies a 3D cross-attention block to strengthen complementary mixed-modality representations. 
Yang et al. \cite{yang2021attention} first obtain the fused feature by concatenating the IRRG and DSM features. 
Then, they concatenate the two weighted fused features using the spatial and channel weights to obtain the final fused feature.
DSHFNet \cite{feng2023dshfnet} builds a spatial self-attention map from one modality, then uses it to reweight the value features of the other two modalities. The final aggregation feature produces enhanced mixed-modality fusion representations.
Studies \cite{zhang2020hybrid, ferrari2021integrating} perform attention-aware multimodal fusion by globally pooling each modality to obtain channel statistics, learning modality-specific channel weights via a gating network, reweighting each modality’s feature channels, and summing the recalibrated features to output a fused representation.
One variant approach is to extract prior features from each modality separately to enhance the principal features, as proposed in \cite{xue2022deep}.

To fully leverage the autonomous learning capabilities of neural networks, many studies also employ pure symmetric network architectures to extract distinct features from each modality.
In Zhang et al. \cite{zhang2021mutual}, a symmetric mutual-guidance attention module generates attention maps in each branch, enabling bidirectional cross-modal supervision that enhances informative structures and suppresses irrelevant responses before multilevel fusion and classification.
MAHiDFNet \cite{wang2022multi} applies identical spatial self-attention on both the HSI spatial and LiDAR branches to refine modality-specific cues before fusion. The same idea can also be seen in MACN \cite{li2023mixing}, CMFAEN \cite{zhang2024cross}, MSFMamba \cite{gao2025msfmamba}, CMFNet \cite{ma2022crossmodal}, MCANet \cite{li2022mcanet} and FTransUNet \cite{ma2024multilevel}.

\begin{table*}[!t]
\caption{\textbf{Summary of unimodal RS datasets used for SS, where SemanticSeg is an abbreviation of semantic segmentation.} }
\label{Dataset1}
\centering
\footnotesize
\begin{tabular}{cllllllll}
\toprule
\textbf{Type} & \textbf{Datasets} & \textbf{Image size} & \textbf{GSD(m)} & \textbf{Classes } & \textbf{Area}(\(km^2\)) & \textbf{Labels} & \textbf{Task}& \textbf{Year} \\
\midrule

\multirow{16}{*}{HSI} 
& Indian Pines (IP)     & 1 × 145 × 145 × 224(200)       & 20      & 16      & 8.41      & 10,249  & Classification &1992 \\
& Washington DC    & 1 × 1208 × 307 × 210(191)      & 1.5-3.0      & 7     & 1.48      & 26,332 & Classification  & 1995 \\
& Kennedy Space Center (KSC)       & 1 × 512 × 614 × 224(176)         & 18           & 13      & 101.86    & 4,756  & Classification  & 1996 \\
& Cuprite           & 1  × 250 × 191  ×    224     & 20          & 30      & 19.1      & 47,750 & Classification  & 1997 \\
& Salinas Valley (SV)    & 1 × 512 × 217 × 224(204)       & 3.7                     & 16      & 1.52      & 54,129  & Classification & 1998 \\
& University of Pavia (UP)           & 1 × 610 × 340 × 115(103)      & 1.3                     & 9       & 0.63      & 42,776 & Classification  & 2001 \\
& Center of Pavia            & 1 × 1096 × 715 × 115(102)          & 1.3       & 9       & 2.03      & 148,152   & Classification  & 2001 \\
& HOSD  \cite{duan2023hyperspectral}           & 18 × Variable × 224     & 3.2-8.1                 & 2       & -         & 14.84M  & SemanticSeg   & 2010 \\
& Hyrank Dioni      & 1 × 250 × 1376 × 176    & 30         & 12      & 309.6     & 20,024  & Classification & 2017 \\
& Hyrank Loukia     & 1 × 249 × 945 × 176     & 30          & 14      & 211.78    & 13,503  & Classification & 2017 \\
& Matiwan Village   & 1 × 3750 × 1580 × 250       & 0.5     & 20      & 1.48      & 5.925M  & Classification & 2017 \\
& WHU-Hi HanChuan \cite{zhong2018mini}   & 1 × 1217 × 303 × 270  & 0.109  & 16 & 0.0044 & 368,751  & Classification  & 2018 \\
& WHU-Hi HongHu \cite{zhong2018mini}    & 1 × 940 × 475 × 270   & 0.043  & 22 & 0.00083 & 446,500  & Classification  & 2018 \\
& WHU-Hi LongKou \cite{zhong2018mini}    & 1 × 550 × 400 × 270  & 0.463  & 9  & 0.047   & 220,000  & Classification  & 2018 \\
& Xiongan \cite{yi2020aerial} &  1 × 3750 × 1580 × 250 &  0.5  &  19  &  1.481  &  2,941,881  & Classification  & 2020 \\
& AeroRIT \cite{rangnekar2020aerorit}  & 1 × 1973 × 3975 × 372  & 0.4    & 5  & 1.25    & 7.843M & SemanticSeg & 2020 \\
& WHU-OHS \cite{li2022whu}       & 7795 × 512 × 512 × 32        & 10     & 24 & 26.21   & 90M   & SemanticSeg  & 2024 \\

\midrule
\multirow{4}{*}{MSI}  
& Zurich Summer     & 20 × 1000 × 1150 ×  4   & 0.61     & 8    & 8.56     & 23M   & SemanticSeg  & 2015 \\
& RIT-18  \cite{kemker2018algorithms}  & 3 × Variable ×  6  &  0.047    & 18   & 0.46  & 209 M  & SemanticSeg  & 2017 \\
& LandCoverNet \cite{DBLP}   & 8941 × 256 × 256 × 10   & 10  & 7   & 58596   & 585.96M   & SemanticSeg & 2020 \\
& MADOS	\cite{KIKAKI202439}   & 6754 × 240 × 240 × 11    &10    &15	&38903	   & 389.03M	& SemanticSeg & 2024 \\

\midrule
\multirow{3}{*}{SAR}  
&OSI \cite{rs11151762}	&1112 × 1250 × 650 × 1	&10	    &5	    &90350	   &903.2M	 & SemanticSeg   &2019 \\
&SOS-G \cite{zhu2021oil}	&3877 × 256 × 256 × 1	&12.5	    &2   	&39700.48	&254.08M	& SemanticSeg  &2022 \\
&SOS-P \cite{zhu2021oil}	&4193 × 256 × 256 × 1	&5  × 20	&2  	&27479.24	&274.79M	& SemanticSeg  &2022 \\

\midrule
\multirow{10}{*}{HRI} 
& SpaceNet1         & 6000 × Variable × 3     & 0.5     & 2       & 2544  & -     & SemanticSeg  & 2017 \\
& SpaceNet2         & 24586 × 650 × 650 × 3   & 0.3     & 2       & 3011  & 10.39B & SemanticSeg & 2017 \\
& INRIA \cite{maggiori2017can}    & 360   × 1500 × 1500 × 3       & 0.3         & 2       & 810       & 810M   & SemanticSeg & 2017 \\
& DeepGlobe  \cite{Demir_2018_CVPR_Workshops}       & 1146 × 2448 × 2448 × 3       & 0.5                     & 7       & 1716.9    & 6.87B   & SemanticSeg & 2018 \\
& Zeebruges    & 7 × 1000 × 1000 × 3    & 0.05     & 8      & 1.75    & 7M     & SemanticSeg  & 2018 \\
& GID  \cite{tong2020land}        & 150 × 6800 × 7200 × 3    & 4      & 5       & 506     & 7.34B  & SemanticSeg & 2020 \\
& SkyScapes   \cite{azimi2019skyscapes}  & 16 × 5616 × 3744 × 3      & 0.13      & 31      & 5.69     & 0.34M  & SemanticSeg & 2019 \\
& GID-Fine   \cite{tong2020land}  & 30000 × 56 × 56 × 3      & 4      & 15      & 506     & 94.08M  & SemanticSeg & 2020 \\
& UAVid  \cite{LYU2020108}          & 300 1.5 Variable × 3       & -      & 8       & -       & 2.5B   & SemanticSeg  & 2020 \\
& LandCover.ai  \cite{Boguszewski_2021_CVPR}    & \begin{tabular}[c]{@{}c@{}}33   × 9000 × 9500 × 3\\      8 × 4200 × 4700 × 3\end{tabular} & 0.25-0.5     & 4   & 216.27   & 2.98B  & SemanticSeg & 2021 \\
& LoveDA \cite{wang2022loveda}           & 5987   × 1024 × 1024 × 3      & 0.3 & 7       & 536.15    & 12B    & SemanticSeg  & 2021 \\
& FloodNet \cite{rahnemoonfar2021floodnet}  & 2343 × 4000 × 3000 × 3 & 0.015  & 9  & 6.3 & 28B  & SemanticSeg &  2021 \\
\bottomrule
\end{tabular}
\end{table*}

\subsubsection{Gated mechanism fusion}

While attention-based fusion techniques are widely used, they typically generate attention weights based on similarity calculations, which incurs excessively high training costs, increasing the computational burden and reducing model efficiency.
Gated mechanism fusion addresses this issue by generating a learnable gating coefficient directly via the network.
The gating coefficient controls the contributions of different features across modalities and determines the optimal intermediate representation.
Based on the operating conditions, gated mechanism fusion methods primarily include independent-gated, complementary-gated, and cross-gated mechanisms, as depicted in Fig. \ref{GatedUnit}.

The independent gate plays an important role in both screening and reconstructing informative features, which uses the independent gate module to control each modality's contribution to the final segmentation.
GRRNet \cite{huang2019automatic} introduced a gated feature selection and fusion system that adaptively integrates both low-level and high-level encoder features into a unified feature map to guide decoding. 
Li et al. \cite{li2023mixing} introduced MACN for multisource RS classification, where the adaptive CNN encoder gates a dual-branch CNN and selects suitable 2D/3D-style convolutions based on input shape, better preserving HSI spectral–spatial structure while extracting LiDAR elevation cues for later fusion.
BCLNet \cite{yue2024bclnet} proposed a gated-attention fusion framework that combines the strengths of gating and attention mechanisms to extract common features and minimize modality discrepancies across heterogeneous RSIs.
However, this isolated fusion strategy lacks information sharing, making it difficult to obtain meaningful complementary features.

The cross-gated mechanism comprises two gate modules, each employing multimodal information transmission between features to achieve pure information exchange.
Kang et al. \cite{kang2022cfnet} proposed a cross-gate fusion module that learns SE-based gates from each modality and cross-controls the other, enabling bidirectional information flow to filter redundant features and exploit complementarity for higher accuracy.
However, the cross-gated fusion methods only perform cross-control of multimodal features at the final stage of the fusion module, making it difficult to adequately filter redundant information.
In contrast, the complementary gate first fuses multimodal images at the input, and then, through a learnable gating function, selects and combines complementary characteristics from both modalities. 
It achieves a balance between the independent gate and the cross gate, realizing a seamless integration of features \cite{li2020collaborative, zhang2023sar, li2024automatic, zhou2021cegfnet, hosseinpour2022cmgfnet, geng2023multisource, wei2024mgfnet}.

The implementation of the complementary gate fusion falls into two categories: the first involves computing a gating coefficient $g$ after merging multimodal inputs, then using $g$ and $1-g$ to calculate the contribution of each modality to the final fusion model \cite{hosseinpour2022cmgfnet, geng2023multisource}.
For example, CHGFNet \cite{li2020collaborative} employed this type of gate for land cover classification, taking into account the varying influence of each modality when classifying different land categories.
Subsequently, Li et al. \cite{li2024automatic} implemented a similar gating mechanism where gate weights were derived from multispectral, SAR and slope data. This improves the reliability of impervious surface mapping in complex topographic regions.
Zhang et al. \cite{zhang2023sar} extract features from original and superpixel-smoothed images in two branches for SAR image classification, then use a gated channel attention module to generate channel-wise gates and adaptively fuse feature maps, balancing misclassification smoothing and detail preservation.

The second approach feeds the fused features into multiple independent gated modules to obtain gating coefficients that control branch modalities.
At each fusion stage, the concatenated features in CEGFNet \cite{zhou2021cegfnet} are fed into two independent gating modules to produce modality-wise coefficients, which reweight each branch before conventional convolutional encoding–decoding in order to learn complementary representations. 
MGFNet \cite{wei2024mgfnet} performs optical–SAR SS using an MLP-Gate fusion module to learn complementary gating coefficients that reweight each modality before feature concatenation, which shows the ability to extract more complex, nonlinear cross-modal dependencies and better suppress redundant or noisy channels.

\subsubsection{Reconstruction mechanism fusion}
Multimodal data show very high heterogeneity. To reduce the redundant information in the fused feature, a feasible solution is to learn a more compact intermediate representation directly. 
Inspired by the AE, the reconstruction mechanism fusion is proposed, which constrains the model by enforcing a reconstruction process during training.
The reconstruction process enhances completeness, reduces redundancy, and improves the discriminative capacity of multimodal middle-feature representations.
They can be roughly divided into self-reconstruction fusion \cite{ding2022global, hong2020deep, zhang2021information, lu2023coupled, zhang2024multimodal, chen2023incomplete} and cross-reconstruction fusion \cite{wu2021convolutional, zhang2018feature, pande2023self}.

Self-reconstruction fusion learns compact feature representation by minimizing the discrepancy between input and reconstructed data. 
EndNet \cite{hong2020deep} used an AE structure to reconstruct raw patch images directly.
GLT Net \cite{ding2022global} and MIViT \cite{zhang2024multimodal} focus on reconstructing encoded multi-scale local spatial features using CNNs with the aid of Transformer modules.
In particular, MIViT employs information aggregation and distribution flows to generate non-redundant, complementary features for classification.
Lu et al. \cite{lu2023coupled} incorporated adversarial learning into a multimodal context, where generators and discriminators engage in an adversarial process.
This approach preserves fine details and complementary information across modalities, allowing for the extraction of high-order semantic features.

The other form of reconstruction approach, cross-reconstruction fusion, transforms one modality into another rather than reconstructing the original input \cite{wu2021convolutional, zhang2018feature, pande2023self}.
At the data level, Zhang et al. \cite{zhang2018feature} reconstructed LiDAR data from HSI within an unsupervised feature extraction framework.
The hidden representation generated during translation preserves feature quality, even when training with limited labeled samples.
At the feature level, CCR Net \cite{wu2021convolutional} first extracts modality-specific features using CNNs and then fuses them through a cross-channel reconstruction module.
Shivam \cite{pande2023self} designed a self-looping CNN that first extracts pre-fusion features and then reconstructs one modality's representation from another, facilitating more discriminative feature learning.

Reconstruction-mechanism fusion is relatively rare in tilewise SS. 
A few works adopt the reconstruction mechanism fusion to maintain performance under missing inputs in small datasets \cite{hong2020multimodal} or modality-incomplete inference \cite{chen2023incomplete}. The main reason is that dense SS pipelines are already compute- and memory-intensive, and adding extra decoders, plus reconstruction objectives, substantially increases optimisation cost and implementation complexity.

\subsubsection{Alignment mechanism fusion}

Alignment mechanism fusion aims to reduce the modality gap before aggregation so that fused representations remain semantically consistent. 
In practice, alignment is less standardized than attention or gated fusion, and it often appears as parameter-sharing constraints, adversarial regularization, or explicit alignment objectives \cite{hong2020more, hong2020multimodal, Hang2022CrossModality, li2023progressive, li2023aligning, gupta2025mosaic, chen2025mdca, li2025semantic}. 
For example, Hong et al. \cite{hong2020more} propose a cross-fusion design that reuses unimodal weights, enabling cross-modal transfer while keeping the backbone compact. 
Hang et al. \cite{Hang2022CrossModality} learn a shared embedding for hyperspectral and LiDAR data via a shared projection and a cross-modality contrastive objective that pulls together representations from the same spatial location. 
For optical–SAR fusion, Li et al. \cite{li2023aligning} introduce a semantic distribution alignment loss based on MMD to map high-level features into a shared latent space. 
Then, MMFNet \cite{li2023progressive} develops a progressive fusion framework that extracts modal-invariant components such as phase features to bridge residual semantic gaps. 
Complementarily, Hong et al. \cite{hong2020multimodal} integrate self-GANs and mutual-GANs to learn perturbation-insensitive representations and explicitly reduce inter-modality discrepancy, yielding more robust cross-modal segmentation.

Recent work further broadens the landscape of alignment mechanism fusion, especially for optical-SAR data where geometric distortions and scale inconsistency make ``align-first, fuse-later'' increasingly common. 
Chen et al. \cite{chen2025mdca} propose MDCA-Net, introducing a cross-modal multi-directional alignment module to mitigate modality discrepancies induced by different imaging mechanisms before subsequent context aggregation and fusion. 
Li et al. \cite{li2025semantic} develop SACFNet for multimodal SS and explicitly position alignment as a prerequisite to fusion by designing a multi-scale, multi-directional alignment module to rectify scale/geometric mismatches between optical and SAR features prior to contextual fusion and dynamic integration. 
Complementarily, MoSAiC \cite{gupta2025mosaic} demonstrates an embedding-level alignment route, where intra- and inter-modality contrastive objectives exploit the natural co-registration of multi-sensor patches to enforce semantic consistency across modalities, yielding more transferable multimodal representations under limited labels. 


\begin{table*}[!t]
\caption{\textbf{Summary of multimodal RS datasets used for SS, where ReferringSeg and ReasoningSeg are abbreviations of referring segmentation and reasoning segmentation, respectively.}}
\label{Dataset2}
\centering
\footnotesize
\begin{tabular}{cllllllll}
\toprule
\textbf{Datasets} & \textbf{Type} & \textbf{Image size} & \textbf{GSD(m)} & \textbf{Classes} & \textbf{Area(\(km^2\))} & \textbf{Labels}& \textbf{Task}  & \textbf{Year} \\
\midrule

\multirow{2}{*}{Trento} &
HSI & 1 × 166 × 600 × 63 & \multirow{2}{*}{1} & \multirow{2}{*}{6} & \multirow{2}{*}{0.1} & \multirow{2}{*}{30,414} & \multirow{2}{*}{Classification} & \multirow{2}{*}{2007} \\
& LiDAR & 1 × 166 × 600 × 2 & & & & & \\
\cmidrule(lr){1-9}

\multirow{2}{*}{Berlin} &
HSI & 1 × 1723 × 476 × 224 & \multirow{2}{*}{30} & \multirow{2}{*}{8} & \multirow{2}{*}{738.13} & \multirow{2}{*}{464,671} & \multirow{2}{*}{Classification} & \multirow{2}{*}{2009} \\
& SAR & 1 × 1723 × 476 × 4 & & & & & \\
\cmidrule(lr){1-9}

\multirow{2}{*}{\makecell{MUUFL \\ Gulfport}} &
HSI & 1 × 325 × 220 × 64(72) & \multirow{2}{*}{1} & \multirow{2}{*}{11} & \multirow{2}{*}{0.07} & \multirow{2}{*}{53,687} & \multirow{2}{*}{Classification} & \multirow{2}{*}{2010} \\
& LiDAR & 1 × 325 × 220 × 2 & & & & & \\
\cmidrule(lr){1-9}

\multirow{2}{*}{DFC 2013} &
HSI & 1 × 1095 × 349  × 144 & \multirow{2}{*}{2.5} & \multirow{2}{*}{15} & \multirow{2}{*}{2.39} & \multirow{2}{*}{15,029} & \multirow{2}{*}{Classification} & \multirow{2}{*}{2012} \\
& LiDAR & 1 × 1095 × 349  × 1 & & & & & \\
\cmidrule(lr){1-9}

\multirow{3}{*}{\makecell{ISPRS \\ Vaihingen}} &
RGB & 33 × Variable × 3 & \multirow{3}{*}{0.09} & \multirow{3}{*}{6} & \multirow{3}{*}{1.34} & \multirow{3}{*}{168M} & \multirow{2}{*}{SemanticSeg} & \multirow{3}{*}{2013} \\
& LiDAR & 33 × Variable × 1 & & & & & \\
& DSM & 33 × Variable × 1 & & & & & \\
\cmidrule(lr){1-9}

\multirow{3}{*}{\makecell{ISPRS \\ Potsdam}} &
MSI & 38 × 6000 × 6000 × 4 & \multirow{3}{*}{0.05} & \multirow{3}{*}{6} & \multirow{3}{*}{3.42} & \multirow{3}{*}{1.37B} & \multirow{2}{*}{SemanticSeg} & \multirow{3}{*}{2013} \\
& LiDAR & 38 × 6000 × 6000 × 1 & & & & & \\
& DSM & 38 × 6000 × 6000 × 1 & & & & & \\
\cmidrule(lr){1-9}

\multirow{3}{*}{DFC 2018} &
HSI & 1 × 601 ×  2384 × 48 & 1 & \multirow{3}{*}{20} & \multirow{3}{*}{1.43} & \multirow{3}{*}{2.02M} & \multirow{2}{*}{Classification}  & \multirow{3}{*}{2017} \\
& LiDAR & 1 × 1202 ×  4768 × 3  & \multirow{2}{*}{0.5} & & & & \\
& RGB & 1 × 1202 ×  4768 × 3 & & & & & \\
\cmidrule(lr){1-9}

\multirow{3}{*}{Augsburg \cite{HONG202168}} &
HSI & 1 × 332 × 485 × 180 & \multirow{3}{*}{30} & \multirow{3}{*}{7} & \multirow{3}{*}{144.92} & \multirow{3}{*}{78,293} & \multirow{2}{*}{Classification} & \multirow{3}{*}{2021} \\
& SAR & 1 × 332 × 485 × 4 & & & & & \\
& LiDAR & 1 × 332 × 485 × 1 & & & & & \\
\cmidrule(lr){1-9}

\multirow{2}{*}{LCZ \cite{hong2020more}} &
MSI & 1 × 626 × 643 × 10 & \multirow{2}{*}{100} & \multirow{2}{*}{10} & \multirow{2}{*}{4025.18} & \multirow{2}{*}{30,087}  & \multirow{2}{*}{Classification}  & \multirow{2}{*}{2021} \\
& SAR & 1 ×  626 × 643 × 4 & & & & & \\
\cmidrule(lr){1-9}

\multirow{5}{*}{\makecell{C2Seg-AB \cite{HONG2023113856}}} &
HSI & \begin{tabular}[c]{@{}c@{}}1 × 2465 × 811 × 242\\      1 × 886 × 1360 × 242\end{tabular} & \multirow{5}{*}{10} & \multirow{5}{*}{13} & \multirow{5}{*}{\makecell{20 \\ 12.05}} & \multirow{5}{*}{\makecell{2M \\ 1.21M}} & \multirow{5}{*}{SemanticSeg} & \multirow{5}{*}{2023} \\
& MSI & \begin{tabular}[c]{@{}c@{}}1 × 2465 × 811 × 4\\    1 × 886 × 1360 × 4\end{tabular} & & & & & \\
& SAR & \begin{tabular}[c]{@{}c@{}}1 × 2465 × 811 × 2\\  1 × 886 × 1360 × 2\end{tabular} & & & & & \\
\cmidrule(lr){1-9}

\multirow{5}{*}{\makecell{C2Seg-BW \cite{HONG2023113856}}} &
HSI & \begin{tabular}[c]{@{}c@{}}1 × 13474 × 8706 ×   116(330)\\      1 × 6225 × 8670 × 116(330)\end{tabular} & \multirow{5}{*}{10} & \multirow{5}{*}{13} & \multirow{5}{*}{\makecell{ 1173.05\\539.71}} & \multirow{5}{*}{\makecell{117.31M\\5397M}} & \multirow{5}{*}{SemanticSeg} & \multicolumn{1}{l}{\multirow{5}{*}{2023}} \\
& MSI & \begin{tabular}[c]{@{}c@{}}1 × 13474 × 8706 × 4\\     1 × 6225 × 8670 × 4\end{tabular} & & & & & \multicolumn{1}{l}{} \\
& SAR & \begin{tabular}[c]{@{}c@{}}1 × 13474 × 8706 × 2\\      1 × 6225 × 8670 × 2\end{tabular} & & & & \multicolumn{1}{l}{} \\
\cmidrule(lr){1-9}

\multirow{4}{*}{MDAS \cite{hu2023mdas}} &
SAR & 1 × 888 × 1371 × 2 & 10 & \multirow{4}{*}{16} & \multirow{4}{*}{121.75} & \multirow{4}{*}{-} &  \multirow{4}{*}{SemanticSeg}  & \multirow{4}{*}{2023} \\
& LiDAR & 1 × 29600 × 45700 × 1 & 0.25 & & & & \\
& MSI & 1 × 888 × 1371 × 12 & 10 & & & & \\
& HSI & 1 × 4036 × 6232 × 368 & 2.2 & & & & \\
\cmidrule(lr){1-9}

\multirow{5}{*}{Ticino \cite{barbato2024ticino}} &
RGB & 1502 × 256 × 362 × 3 & 1.86-2.64 & \multirow{5}{*}{8/10} & \multirow{5}{*}{1331.72} & \multirow{5}{*}{-} &  \multirow{4}{*}{SemanticSeg}  & \multicolumn{1}{l}{\multirow{5}{*}{2024}} \\
& PAN & 1502 × 96 × 192 × 1 & \multirow{4}{*}{5} & & & & \\
& HSI VNIR & 1502 × 96 × 192  ×   60(63) & & & & & \\
& HSI SWIR & 1502 × 96 × 192 × 122(171) & & & & & \\
& DTM & 1502 × 101 × 203 × 1 & & & & & \\
\cmidrule(lr){1-9}

\multirow{3}{*}{SZUTreeData-R1 \cite{li2023texture}} &
RGB & 1 × 6170 × 4810 × 3 & 0.05 & \multirow{3}{*}{20} & \multirow{3}{*}{0.07} & \multirow{3}{*}{492,631} & \multicolumn{1}{l}{\multirow{3}{*}{Classification}} & \multicolumn{1}{l}{\multirow{3}{*}{2025}} \\
& HSI & 1 × 3085 × 2405 × 112 & \multirow{2}{*}{0.1} & & & & \\
& LiDAR & 1 × 3085 × 2405 × 1 & & & & & \\
\cmidrule(lr){1-9}

\multirow{3}{*}{SZUTreeData-R2 \cite{li2023texture}} &
RGB & 1 × 8080 × 4888 × 3 & 0.05 & \multirow{3}{*}{21} & \multirow{3}{*}{0.1} & \multirow{3}{*}{696,620}  & \multicolumn{1}{l}{\multirow{3}{*}{Classification}} & \multicolumn{1}{l}{\multirow{3}{*}{2025}} \\
& HSI & 1 × 4040 × 2444 × 112 & \multirow{2}{*}{0.1} & & & & \\
& LiDAR & 1 × 4040 × 2444 × 1 & & & & & \\
\cmidrule(lr){1-9}

\multirow{2}{*}{RefSegRS \cite{yuan2024rrsis}} &
Text & 4420 & \multirow{2}{*}{0.13} & \multirow{2}{*}{14} & \multirow{2}{*}{19.58} & \multirow{2}{*}{-} & \multirow{2}{*}{ReferringSeg} & \multirow{2}{*}{2024} \\
& RGB &  4420 × 512 × 512 × 3 & & & & & \\
\cmidrule(lr){1-9}

\multirow{2}{*}{RRSIS-D \cite{liu2024rotated}} &
Text & 17402 & \multirow{2}{*}{0.5-30} & \multirow{2}{*}{20} & \multirow{2}{*}{-} & \multirow{2}{*}{-}  & \multirow{2}{*}{ReferringSeg}  & \multirow{2}{*}{2024} \\
& RGB &  17402 × 800 × 800 × 3 & & & & & \\
\cmidrule(lr){1-9}

\multirow{2}{*}{RISBench \cite{dong2024cross}} &
Text & 52472 & \multirow{2}{*}{0.1-30} & \multirow{2}{*}{26} & \multirow{2}{*}{-} & \multirow{2}{*}{-} & \multirow{2}{*}{ReferringSeg}  & \multirow{2}{*}{2025} \\
& RGB &  52472 × 512 × 512 × 3 & & & & & \\
\cmidrule(lr){1-9}

\multirow{2}{*}{EarthReason \cite{li2025segearth}} &
Text & 5434 & \multirow{2}{*}{0.5-153} & \multirow{2}{*}{28} & \multirow{2}{*}{-} & \multirow{2}{*}{-}  & \multirow{2}{*}{ReasoningSeg}  & \multirow{2}{*}{2025} \\
& RGB &  5434 × Variable × 3 & & & & & \\

\bottomrule
\end{tabular}
\end{table*}

\section{Datasets for RSISS}
\label{Datasets}

As one of the most advanced forms of data-driven modeling, DL models, particularly those based on neural networks, require large volumes of data to automatically learn patterns, features, and relationships without manual feature engineering.
The development and performance of these models are highly dependent on both the quality and quantity of training data.

With the increasing availability of airborne and spaceborne sensors, high-quality RSIs are now accessible on a daily basis.
To facilitate efficient identification of relevant datasets, a list of the most representative datasets for RSISS is compiled, covering a wide range of thematic areas.
As depicted in Tables~\ref{Dataset1} and~\ref{Dataset2}, these RS datasets support the advancement of RSISS by enabling more comprehensive training and evaluation across diverse tasks.
RSIs differ from conventional street-view imagery.
Captured from high altitudes, this ``bird's-eye view'' allows for broad area observation but introduces unique challenges in data processing \cite{hong2020multimodal}.
In addition, dataset selection for the four training paradigms relies on empirical guidelines dictated by data characteristics. Massive spatial datasets require models at the tile- or image-level to accommodate memory limits, whereas smaller or highly spectral datasets suit models at the patch-level to extract local signatures without overfitting.
We list the most common tasks for these datasets.

\subsection{High-resolution datasets}

High-resolution images (HRIs) typically operate within the visible wavelength range, aligning with human visual perception and thereby simplifying the labeling process.
With spatial resolutions at the metre or sub-metre level, HRIs provide rich spatial detail. And, they usually contain only three spectral bands, resulting in smaller data volumes that are easier to store and process.
Due to these advantages, HRIs constitute the largest proportion of labelled RS data.
For example, the SpaceNet initiative released a large-scale dataset comprising over 3000 km\textsuperscript{2} of coverage and more than 10 billion labelled objects.
Similarly, LoveDA \cite{wang2022loveda} offers over 12~billion labelled instances across three cities.
Such large-scale benchmark datasets enable the development of models with strong generalisation capability.

\subsection{Hyperspectral datasets}

HSIs capture information across a broad range of wavelengths, extending far beyond the visible spectrum.
Each pixel in an HSI corresponds to a spectral vector that describes the interaction of light with materials at multiple wavelengths, enabling precise material identification and signal recognition \cite{wang2023frequency}.
Since the 1990s, datasets such as Indian Pines and Washington DC have been widely used in land cover and land use studies.
However, the high dimensionality of HSI leads to large file sizes and complex processing, limiting the quantity and diversity of publicly available hyperspectral datasets \cite{wang2024sliding}.
Recently, WHU-OHS \cite{li2022whu} was introduced as the first large-scale, publicly available hyperspectral dataset.
It contains approximately 90~million manually labelled samples, with extensive geographic distribution and broad spatial coverage, offering new opportunities for robust model development and benchmarking.

\subsection{Multispectral datasets}

Multispectral images (MSIs) provide a balance between HSIs and HRIs by offering high-quality spatial and spectral information at a smaller data scale.
This makes them more efficient to process while still capturing essential spectral features for RS applications.
The Sentinel-2 mission has markedly contributed to the availability of free MSIs for the RS community.
These data support a wide range of applications, including agricultural monitoring, emergency management, water quality assessment, and LULC analysis.
For instance, LandCoverNet \cite{DBLP} offers a global benchmark dataset for land cover classification using Sentinel-2 imagery at 10~m spatial resolution.

\subsection{SAR datasets}

As an active RS technique, SAR operates independently of lighting and most weather conditions, enabling continuous day-and-night imaging.
SAR is highly sensitive to changes in surface roughness, making it particularly effective for detecting phenomena such as wind speeds, wave heights, ocean eddies, and surface composition.
Due to these capabilities, SAR systems have become essential tools for ocean observation.
Zhu et al. introduced a manually labeled dataset focused on oil spill detection in the Gulf of Mexico and Persian Gulf regions.
Marios et al. \cite{rs11151762} compiled oil spill data from Sentinel-1 observations across Europe between 2015 and 2017, offering a valuable benchmark for future SAR-based oil spill detection research.

\subsection{LiDAR datasets}

LiDAR provides precise 3-D spatial information, making it critical for applications requiring detailed topographic and structural analysis.
Using laser pulses to measure distances, LiDAR systems generate dense point clouds that represent both ground surfaces and above-ground features, such as buildings and vegetation.
A key advantage of LiDAR is its ability to penetrate vegetation canopies, enabling accurate ground elevation mapping.
However, LiDAR typically involves large file sizes and requires complicated preprocessing, including filtering and interpolation, to produce derived products such as digital terrain models (DTMs) and canopy height models.
Despite these challenges, LiDAR datasets remain indispensable in urban planning, forestry monitoring, and disaster response, offering high-resolution, scalable insights across diverse application domains.

\subsection{Multimodal datasets}

Multimodal RS combines data from different sensor types to extract complementary information, resulting in more comprehensive and accurate surface understanding and improved performance in downstream tasks.
Early datasets such as Trento and Berlin were used to evaluate multimodal fusion performance, but were limited to two modalities and contained minimal annotation.
With the introduction of MDAS \cite{hu2023mdas} and Ticino \cite{barbato2024ticino}, multimodal RS benchmark datasets have become more diverse and extensive.
Ticino includes five modalities: RGB, DTMs, panchromatic, HSI in the visual–near-infrared (VNIR) range, and HSI in the short-wave infrared (SWIR) range.
This dataset offers a robust benchmark for evaluating multimodal fusion algorithms across complex and heterogeneous inputs.
More recently, natural language text has been introduced as a novel modality to align visual features with semantic concepts, enabling advanced vision-language tasks in the remote sensing domain. 
Emerging large-scale datasets, such as RRSIS-D \cite{liu2024rotated}, EarthReason \cite{li2025segearth}, pair HRIs with rich textual descriptions to support complex multimodal applications like referring segmentation and reasoning segmentation.

\section{Critical Thinking and Future Developments}
\label{OIFD}

In previous sections, we examined the evolution of RSISS through a structural lens: the "pixel–patch–tile–image" hierarchy. 
This granularity-based taxonomy not only reveals progress in end-to-end SS but also exposes the core engineering bottlenecks inherent in each scale. 
While conventional reviews often attribute future breakthroughs to "deeper networks" or "larger datasets," this survey posits that the future of RSISS depends fundamentally on how researchers navigate the trade-offs among segmentation granularities and, ultimately, on how they dismantle the boundaries that separate them. 
Accordingly, we propose critical perspectives and future directions across four key dimensions.

\subsection{The Enduring Value of Pixel/Patch-Level Methods in Low-Label and High-Spectral Regimes}

With the rapid rise of tile-level and image-level models, pixel- and patch-based methods are frequently criticized for their restricted receptive fields and computational redundancy, often being dismissed as obsolete paradigms. 
However, this perspective overlooks the complex multimodal nature of RS data. When processing HSI, the exceedingly high spectral dimensionality makes large-scale spatial modeling susceptible to the curse of dimensionality and severe overfitting. 
In such scenarios, the ``local spatial-spectral focus" maintained by patch-level methods remains one of the most robust feature extraction strategies. 
Furthermore, in specialized domains such as environmental and ecological studies, annotation costs are prohibitively high. Researchers often only have access to sparse, pixel-wise labels, rendering data-hungry tile or image-level models impractical for direct application.

Future research at the pixel and patch level should pivot away from only chasing state-of-the-art metrics on standard optical benchmarks and instead focus on extreme operational conditions:
First, researchers should explore lightweight patch models for multimodal data. This involves investigating how multiple data sources can be fused with hyperspectral patches at a localized level, maximizing local discriminative power while maintaining a highly efficient minimal receptive field. 
Second, investigating active learning in extreme few-shot scenarios is crucial. This requires studying how patch-level uncertainty quantification can guide active learning frameworks to achieve high-precision semantic extraction with minimal human annotation.

\subsection{Bottlenecks and Evolution of Tile-Level Paradigms for Efficient Dense Mapping}

The tile-level paradigm is currently the dominant compromise in both academia and industry for processing HRIs, striking a viable balance between local detail retention and GPU memory constraints. 
However, its fundamental contradiction lies in the fact that artificial physical cropping disrupts genuine geographical topology. 
Edge artifacts at tile seams and the forced truncation of large-scale continuous entities remain the inherent vulnerability of tile-level models, particularly problematic in tasks like large-scale impervious surface extraction.
While many studies attempt to mitigate this via overlapping inference, this introduces multiplied computational redundancy, directly contradicting the overarching goal of highly efficient, large-scale dense mapping.

To realize true large-scale RS information extraction, future tile-level architectures must transcend the limitations of conventional sliding windows. 
First, there is a need for seamless cross-tile contextual interaction. This involves developing architectures that can share boundary features or cache global context in real time during sequential tile inference, eliminating receptive-field truncation at the algorithmic level without relying on physical overlap. 
Second, optimizing efficient dense mapping in resource-constrained environments is essential. For edge computing devices such as airborne platforms, enhancing the inference efficiency of tile-level models is critical to enable high-precision, seamless mapping while avoiding redundant inference.

\subsection{Generalization Limits and Universal Interfaces of Image-Level Foundation Models}

The holistic learning architecture introduces spatial parallel interaction mechanisms, breaking the previous physical limitations of having to crop large images into tiles, and achieves true end-to-end image-level segmentation. 
This paradigm resolves the performance degradation typical of traditional cropping and fusion processes.
Nevertheless, this holistic nature places extreme demands on the scalability of the computing architecture.
When facing ultra-large-scale, ultra-high-resolution, and multimodal heterogeneous imagery, maintaining the highly efficient interaction and global consistency of end-to-end features without degrading to tile processing is currently the greatest challenge at this level. 
Additionally, generalizing this holistic capability from specific tasks to universal geospatial understanding interfaces requires deeper exploration.

The future of image-level models lies in cost-effective adaptation to RS-specific tasks. 
First, exploring PEFT for holistic architectures is paramount. 
Applying techniques like low-rank adaptation to holistic models can enable them to adapt to different sensor parameters and geographical regions with minimal computational overhead. 
Second, constructing multimodal universal holistic interfaces is necessary. Future frameworks should uniformly process optical, synthetic aperture radar, and light detection and ranging data through universal prompt or instruction interfaces to achieve cross-modality end-to-end holistic parsing.

\subsection{Breaking the Hierarchy: The Future of Cross-Regime Transfer}

Reflecting on the four-level taxonomy proposed in this survey, most current research isolates optimization within a single granularity level, such as optimizing a tile model or a global model exclusively. 
However, in practical applications, this segregation creates an awkward situation where global models are too heavy to deploy, and local models are too light to capture the big picture. 
This hierarchical gap hinders the deployment of advanced algorithms in practical engineering tasks like real-time environmental monitoring.

Breaking the boundaries of the ``pixel–patch–tile–image" hierarchy to achieve cross-granularity knowledge transfer represents the most promising frontier. 
First, utilizing cross-granularity knowledge distillation is a key direction. Massive holistic learning models can serve as teacher networks to extract global topological structures and geographical association information, guiding the training of lightweight tile-level student networks. 
Second, this paradigm will facilitate cross-hierarchical consistency modeling. This ensures that even when inference is restricted to local tiles, the model retains the macro perspective learned from the global model, thereby obtaining segmentation results with global consistency while maintaining high inference efficiency.

\section{Conclusion}
\label{Conclusion}

This survey presents a structured review of RSISS in the DL era through a unified pixel–patch–tile–image taxonomy aligned with segmentation granularity and the training/inference pipeline.
Under this operational hierarchy, the progression from pixel- and patch-level classification to tile- and image-level modeling is systematically examined.
By analyzing supervision paradigms, feature extraction strategies, and multimodal fusion mechanisms, the co-evolution of data scale, architectural design, and representation learning is clarified.

The evolution of RSISS reflects a fundamental shift from local discriminative learning toward globally contextualized and data-driven representation learning.
As receptive fields expand and model capacity increases, progress is increasingly governed by scalable training strategies, domain robustness, and multimodal integration rather than architectural complexity alone.
The transition from convolutional models to transformer-based architectures and vision foundation models marks a movement toward more transferable and generalizable representations.

Despite substantial advances in common land-cover segmentation, challenges remain in annotation scarcity, modality heterogeneity, computational scalability, and cross-domain reliability.
Future progress is expected to depend on harmonizing large-scale pretraining with task-specific adaptation, strengthening domain generalization, and developing unified multimodal frameworks for diverse sensing platforms.
%

\bibliographystyle{IEEEtran}
\bibliography{references}

\begin{IEEEbiography}[{\includegraphics[width=1in, height=1.25in, clip, keepaspectratio]{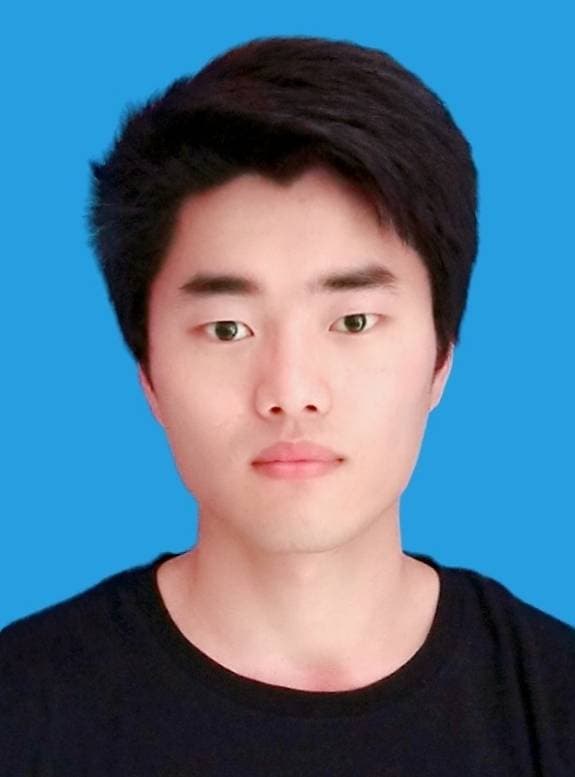}}]{Quanwei Liu} (Student Member, IEEE) received the B.S. degree in exploration technology and engineering from the Henan University of Engineering, Zhengzhou, China, in 2020, and the M.S. degree in resources and environment from the China University of Geosciences, Wuhan, China, in 2023. He is currently pursuing a Ph.D. in multimodal image fusion, hyperspectral image classification, and remote-sensing image segmentation at the College of Science and Engineering and the Centre for AI and Data Science Innovation, James Cook University, Cairns, QLD, Australia.

His research interests include remote image segmentation, hyperspectral image processing, multimodal remote sensing fusion, and deep learning.
\end{IEEEbiography}

\begin{IEEEbiography}[{\includegraphics[width=1in, height=1.25in, clip,keepaspectratio]{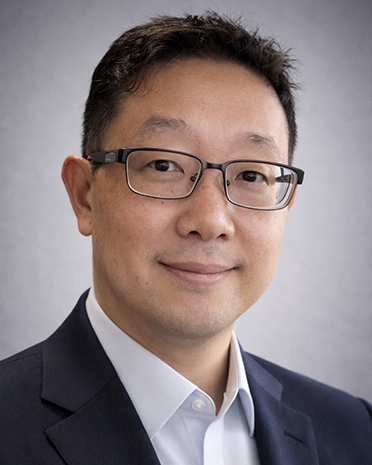}}]
{Tao Huang} (Senior Member, IEEE) received the B.Eng. degree in electronics and information engineering from the Huazhong University of Science and Technology, Wuhan, China, in 2003, the M.Eng. degree in sensor system signal processing from The University of Adelaide, Adelaide, SA, Australia, in 2007, and the Ph.D. degree in electrical engineering from the University of New South Wales, Sydney, NSW, Australia, in 2016. He was an Endeavor Australia Cheung Kong Research Fellow. He is currently a Senior Lecturer at James Cook University (JCU) in Cairns, QLD, Australia, where he serves as a Co-Deputy Director of the JCU Center for AI and Data Science Innovation (CADSI).
Dr. Huang has received several awards, including the Australian Postgraduate Award, the Engineering Research Award from the University of New South Wales, the Best Paper Award at the IEEE Wireless Communications and Networking Conference (2011), the IEEE Outstanding Leadership Award (2022), the IEEE Access Outstanding Associate Editor Recognition (2023, 2024, and 2025),  two Springer Nature Editor of Distinction Awards 2026 (Editorial Contribution Award and Author Service Award), and the JCU Citation for Outstanding Contribution to Student Learning (2022). He currently serves as Vice Chair of the IEEE Northern Australia Section and Chair of the local MTT-S/ComSoc Chapter. He has also held leadership roles at international conferences, including General Co-Chair, Publicity Co-Chair, Workshop Co-Chair, Publication Co-Chair, Technical Program Committee Chair, Program Vice Chair, and Symposium Chair. He is an Associate Editor of Scientific Reports, IEEE Transactions on Vehicular Technology, IEEE Transactions on Emerging Topics in Computing, IEEE Open Journal of the Communications Society, IEEE Access, Frontiers in Artificial Intelligence, and IET Communications.
\end{IEEEbiography}

\begin{IEEEbiography}[{\includegraphics[width=1in, height=1.25in, clip, keepaspectratio]{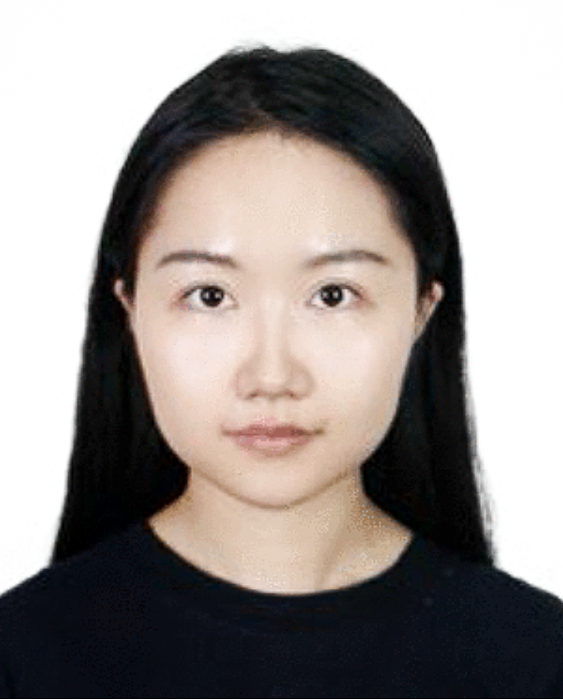}}]{Jiaqi Yang} (Member, IEEE) received the B.S. degree in Geographic Information Science from Dalian Maritime University, Dalian, China, in 2019, the M.E. degree in Geomatics Engineering from the State Key Laboratory of Information Engineering in Surveying, Mapping and Remote Sensing, Wuhan University, Wuhan, China, in 2021, and the Ph.D. degree in Photogrammetry and Remote Sensing from Wuhan University, Wuhan, China, in 2024. 
He is currently a postdoctoral research associate in University of Wisconsin-Madison, Madison, WI USA. 

Her research interests include deep learning, satellite image processing, hyperspectral image interpretation and multimodal data analysis.
\end{IEEEbiography}

\begin{IEEEbiography}[{\includegraphics[width=1in, height=1.25in, clip,keepaspectratio]{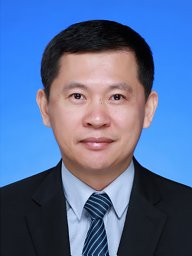}}]
{Wei Xiang} (S’00–M’04–SM’10) is a La Trobe Distinguished Professor and Cisco Research Chair of AI and IoT at La Trobe University. He founded two world-first research centres, namely the CiscoLa Trobe Centre for AI and IoT in 2020, and the Australian Centre for AI and Medical Innovation (ACAMI) in 2024. Previously, he was Foundation Chair and Head of Discipline of IoT Engineering at James Cook University, Cairns, Australia. For the past six consecutive years (2020-2025), has consistently ranked among Stanford University World’s Top 2\% Scientists for both his single-year and career-long impacts. Due to his instrumental leadership in establishing Australia’s first accredited Internet of Things Engineering degree program, he was inducted into the Pearcy Foundation’s Hall of Fame in October 2018. He is a TEDx speaker and an elected Fellow of the IET in UK and Engineers Australia. He received the TNQ Innovation Award in 2016, the Pearcey Entrepreneurship Award in 2017, and the Engineers Australia Cairns Engineer of the Year in 2017. He was a corecipient of four Best Paper Awards at WiSATS’2019, WCSP’2015, IEEE WCNC’2011, and ICWMC’2009. He has been awarded several prestigious fellowship titles. He was named a Queensland International Fellow (2010-2011) by the Queensland Government of Australia, an Endeavour Research Fellow (2012-2013) by the Commonwealth Government of Australia, a Smart Futures Fellow (2012-2015) by the Queensland Government of Australia, and a JSPS Invitational Fellow jointly by the Australian Academy of Science and Japanese Society for Promotion of Science (2014-2015). He was the Vice Chair of the IEEE Northern Australia Section from 2016-2020. He is currently an Associate Editor for IEEE COMMUNICATIONS SURVEYS \& TUTORIALS, IEEE TRANSACTIONS ON VEHICULAR TECHNOLOGY, IEEE INTERNET OF THINGS JOURNAL, IEEE ACCESS, and Nature journal of Scientific Reports. He has published over 450 peer-reviewed papers, including 3 books and nearly 300 journal articles. He has served in a large number of international conferences in the capacity of General Co-Chair, TPC Co-Chair, Symposium Chair, etc. His research interest includes the Internet of Things, wireless communications, machine learning for IoT data analytics, and computer vision.
\end{IEEEbiography}

\EOD

\end{document}